\documentclass{article}

\usepackage{microtype}
\usepackage{graphicx}
\usepackage{booktabs} 
\usepackage{amsfonts}       
\usepackage{nicefrac}       
\usepackage{microtype}      
\usepackage{setspace}
\usepackage{amssymb, amsmath}
\usepackage{bm}
\usepackage{graphicx}
\usepackage{float}
\usepackage{color,soul,xcolor}
\usepackage{afterpage}
\usepackage{subcaption}
\usepackage{makecell}
\usepackage{adjustbox}
\usepackage{threeparttable}
\usepackage{enumitem}

\usepackage{hyperref}

\usepackage[accepted]{icml2021}

\icmltitlerunning{SketchEmbedNet: Learning Novel Concepts by Imitating Drawings}

\begin{document}

\twocolumn[
\icmltitle{SketchEmbedNet: Learning Novel Concepts by Imitating Drawings}

\icmlsetsymbol{equal}{*}

\begin{icmlauthorlist}
\icmlauthor{Alexander Wang}{equal,uoft,vector}
\icmlauthor{Mengye Ren}{equal,uoft,vector}
\icmlauthor{Richard S. Zemel}{uoft,vector,cifar}
\end{icmlauthorlist}

\icmlaffiliation{uoft}{University of Toronto, Toronto, Canada}
\icmlaffiliation{vector}{Vector Institute}
\icmlaffiliation{cifar}{CIFAR}

\icmlcorrespondingauthor{Alexander Wang}{alexw@cs.toronto.edu}
\icmlcorrespondingauthor{Mengye Ren}{mren@cs.toronto.edu}
\icmlcorrespondingauthor{Richard S. Zemel}{zemel@cs.toronto.edu}
\icmlkeywords{Machine Learning, ICML}
\vskip 0.3in
]
\printAffiliationsAndNotice{\icmlEqualContribution} 

\newcommand{\MR}[1]{{\color{orange}MR: #1}}
\newcommand{\RZ}[1]{{\color{magenta}RZ: #1}}
\newcommand{\AW}[1]{{\color{cyan}AW: #1}}
\newcommand{\AWC}[1]{{\color{cyan}#1}}
\newcommand{\model}[0]{SketchEmbedNet}
\newcommand{\modelembedding}[0]{SketchEmbedding}

\begin{abstract}
Sketch drawings capture the salient information of visual concepts. Previous 
work has shown that neural networks are capable of producing sketches of 
natural objects drawn from a small number of classes. While earlier 
approaches focus on generation quality or retrieval, we explore properties 
of image representations learned by training a model to produce sketches of 
images. We show that this generative, class-agnostic model produces 
informative embeddings of images from novel examples, classes, and even 
novel datasets in a few-shot setting. Additionally, we find that these 
learned representations exhibit interesting structure and compositionality.
\end{abstract}


\section{Introduction}
\label{sec:introduction}
Drawings are frequently used to facilitate the communication of new ideas. If someone asked what an apple is, or looks like, a natural approach would be to provide a simple, pencil and paper drawing; perhaps a circle with divots on the top and bottom and a small rectangle for a stem. These sketches constitute an intuitive and succinct way to communicate concepts through a prototypical, visual representation. This phenomenon is also preserved in logographic writing systems such as Chinese hanzi and Egyptian hieroglyphs where each character is essentially a sketch of the object it represents. Frequently, humans are able to communicate complex ideas in a few simple strokes.


Inspired by this 
idea that sketches capture salient
aspects of concepts, 
we hypothesize that it is possible to learn informative representations by expressing them as sketches. In this paper we target the image domain and seek to develop representations of images from which sketch drawings can be generated.
Recent research has explored a wide variety of sketch generation models, ranging from generative adversarial networks (GANs)~\citep{pix2pix, photosketch}, to autoregressive~\citep{draw, ha2017sketchrnn, chen2017pix2seq}, transformer~\citep{sketchformer, aksan2020cose}, hierarchical Bayesian~\citep{Lake2015bpl} and neuro-symbolic~\citep{tian2020learning} models. These methods may generate in pixel-space or in a sequential setting such as a motor program detailing pen movements over a drawing canvas. Many of them face shortcomings with respect to representation learning on images: hierarchical Bayesian models scale poorly, others only generate a single or a few classes at a time, and many require sequential inputs, which limit their use outside of creative applications.

We develop \model{}, a class-agnostic encoder-decoder model that produces a 
``\modelembedding{}'' of an input image as an encoding which is then decoded as a sequential motor program. By knowing ``how to sketch an image,'' it learns an informative representation that leads to strong 
performance on classification tasks despite being learned without class labels. Additionally, training on a broad collection of classes enables strong generalization and produces a class-agnostic embedding function. We demonstrate these claims by showing that our approach generalizes to novel examples, classes, and datasets, most notably in a challenging unsupervised few-shot classification setting on the Omniglot~\citep{Lake2015bpl} and mini-ImageNet~\citep{vinyals2016matching} benchmarks.

While pixel-based methods produce good visual results, they may lack clear component-level awareness, or understanding of the spatial relationships between them in an image; we have seen this collapse of repeated components in GAN literature~\citep{gantutorial}. By incorporating specific pen movements and the contiguous definition of visual components as points through time, \modelembedding{}s encode a unique visual understanding not present in pixel-based methods. We study the presence of componential and spatial understanding in our experiments and also present a surprising phenomenon of conceptual composition where concepts can be added and subtracted through embeddings. 

\begin{figure*}[t]
    \centering
    \includegraphics[width=0.98\textwidth]{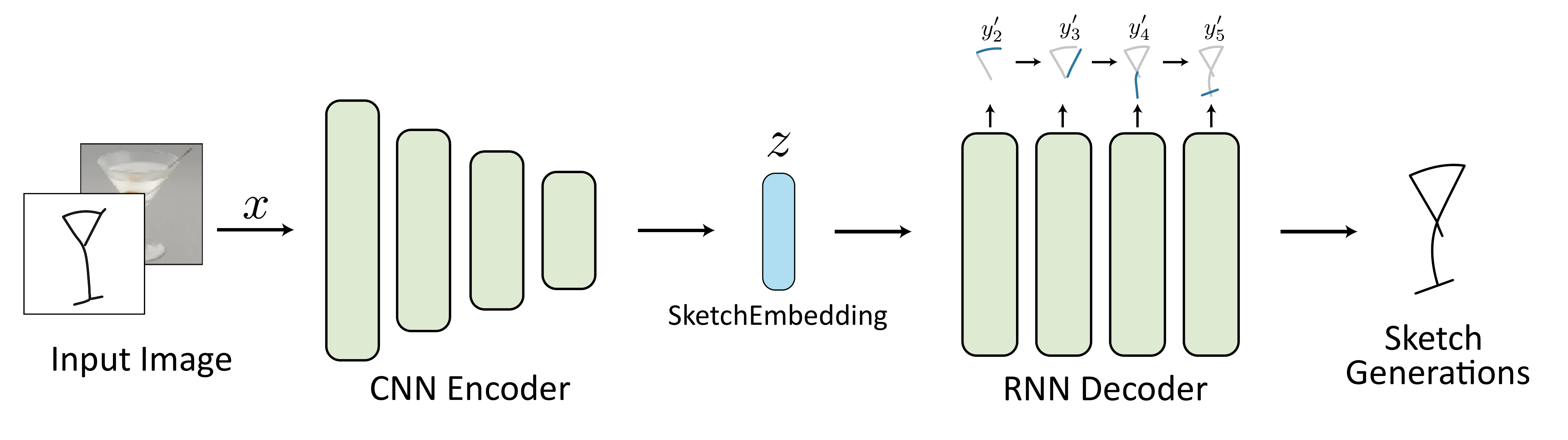}
    \caption[]{\textbf{Representation learning and generation 
    setup} -- a pixel image is encoded as \modelembedding{} $z$ then decoded as a sequential sketch. Model input can be either a sketch image or a natural image.}
    \label{fig:1_hlsetup}
\end{figure*}

\section{Related Work}
\label{sec:related_work}
\paragraph{Sketch-based visual understanding.}
Recent research motivates the use of sketches to understand and classify images. Work by~\cite{Hertzmann2020WhyDL} demonstrated that line drawings are an informative depiction of shape and are intuitive to human perception. \citet{Lamb_2020_WACV} further, proposed that sketches are a detail-invariant representation of objects in an image that summarize salient visual information. \citet{geirhos2018shapebias} demonstrates that a shape-biased perception, is more robust and reminiscent of human perception. We build on this intuition to sketches for shape-biased perception by building a generative model to capture it in a latent representation.

\paragraph{Sequential models for sketch generation.}
Many works study the generation of sequential sketches without specifying individual pixel values; ~\citet{handwrittenmotor} trained a generative model for MNIST~\citep{mnist} examples by specifying spring stiffness to move a pen in 2D space. ~\citet{gravesrnn} introduced the use of an LSTM~\citep{hochreiter1997lstm} to model handwriting as a sequence of points using recurrent networks. SketchRNN~\citep{ha2017sketchrnn} extended the use of RNNs to sketching models that draw a single class. ~\citet{learn2sketchcycle,chen2017pix2seq,sketchformer} made use of pixel inputs and consider more than one class while ~\citet{sketchformer,aksan2020cose} introduced a transformer~\citep{transformers} architecture to model sketches.~\citet{Lake2015bpl} used a symbolic, hierarchical Bayesian model to generate Omniglot~\citep{Lake2015bpl} examples while ~\citet{tian2020learning} used a neuro-symbolic model for concept abstraction through sketching. ~\citet{carlier2020deepsvg} explored the sequential generation of scalable vector graphics (SVG) images. We leverage the SketchRNN decoder for autoregressive sketch generation, but extend it to hundreds of classes with the focus of learning meaningful image representations. Our model is reminiscent of ~\citep{chen2017pix2seq} but to our knowledge no existing works have learned a class-agnostic sketching model using pixel image inputs.

\paragraph{Pixel-based drawing models.} 
Sketches and other drawing-like images can be specified directly in pixel space by outputting pixel intensity values. They were proposed as a method to learn general visual representation in the early literature of computer vision~\citep{Marr}. Since then pixel-based ``sketch images'' can be generated through style transfer and low-level processing techniques such as edge detection~\citep{ucm}. Deep generative models~\citep{pix2pix} using the GAN~\citep{gan} architecture have performed image-sketch domain translation and Photosketch~\citep{photosketch} focused specifically on the task with an $1:N$ image:sketch pairing. ~\citet{Liu2020neural} generates sketch images using varying lighting and camera perspectives combined with 3D mesh information. ~\citet{fcr} used a CNN model to generate sketch-like images of faces. DRAW~\citep{draw} autoregressively generates sketches in pixel space by using visual attention. ~\citet{pixelcnn, oneshotgen} autoregressively generate pixel drawings. In contrast to pixel-based approaches, \model{} does not directly specify pixel intensity and instead produces a sequence of strokes that can be directly rendered into a pixel image. We find that grouping pixels as ``strokes'' improves the object awareness of our embeddings.

\paragraph{Representation learning using generative models.}
Frequently, generative models have been used as a method of learning useful representations for downstream tasks of interest.
In addition to being one of the first sketch-generation works, ~\citet{handwrittenmotor} also used the inferred motor program to classify MNIST examples without class labels. Many generative models are used for representation learning via an \textit{analysis-by-synthesis} approach, e.g., deep and variational autoencoders~\citep{sae, vae}, Helmholtz Machines~\citep{dayan1995helmholtz}, BiGAN~\citep{donahue2016bigan}, etc. Some of these methods seek to learn better representations by predicting additional properties in a supervised manner. Instead of including these additional tasks alongside pixel-based reconstruction, we generate in the sketch domain to learn our shape-biased representations.

\paragraph{Sketch-based image retrieval (SBIR).} SBIR also seeks to map sketches and sketch images to image space. The area is split into fine-grained (FG-SBIR)~\citep{sketchshoe, Sangkloy2016sketchy, sketchlessformore} and a zero-shot setting (ZS-SBIR)~\citep{cyclesbir, stackedsbir, doodletosketch}. FG-SBIR considers minute details, while ZS-SBIR learns high-level cross-domain semantics and a joint latent space to perform retrieval.

\section{Learning to Imitate Drawings}
\label{sec:method}
We present a generative sketching model that outputs a sequential motor program ``sketch'' describing pen movements, given only an input image. It uses a CNN-encoder and an RNN-decoder trained using our novel pixel-loss curricula in addition to the objectives introduced in SketchRNN~\citep{ha2017sketchrnn}.

\subsection{Data representation}
\model{} is trained using image-sketch pairs $(\bm{x}, \bm{y})$, where $\bm{x} \in \mathbb{R}^{H\times W \times C}$ is the input image and $\bm{y} \in \mathbb{R}^{T \times 5}$ is the motor-program representing a sketch. We adopt the same representation of $\bm{y}$ as used in SketchRNN~\citep{ha2017sketchrnn}. $T$ is the maximum sequence length of the sketch data $\bm{y}$, and each "stroke" $\bm{y}_t$ is a pen movement that is described by 5 elements, $(\Delta_x, \Delta_y, s_1, s_2, s_3)$. The first 2 elements are horizontal and vertical displacements on the drawing canvas from the endpoint of the previous stroke. The latter 3 elements are mutually exclusive pen states: $s_1$ indicates the pen is on paper for the next stroke, $s_2$ indicates the pen is lifted, and $s_3$ indicates the sketch sequence has ended. The first "stroke" $\bm{y}_0$ is initialized as (0, 0, 1, 0, 0) for autoregressive generation. Note that no class information is ever provided to the model while learning to draw.

\subsection{Convolutional image embeddings}
We use a CNN to encode the input image $\bm{x}$ and obtain the latent space representation $\bm{z}$, as shown in Figure~\ref{fig:1_hlsetup}. To model intra-class variance, $\bm{z}$ is a Gaussian random variable parameterized by CNN outputs $\mu$ and $\sigma$ like in a VAE~\citep{vae}. Throughout this paper, we refer to $\bm{z}$ as the {\it \modelembedding{}}.

\subsection{Autoregressive decoding of sketches}
The RNN decoder used in \model{} is the same as in SketchRNN~\citep{ha2017sketchrnn}. The decoder outputs a mixture density representing the distribution of the pen offsets at each timestep. It is a mixture of $M$ bivariate Gaussians denoting the spatial offsets as well as the probability over the three pen states $s_{1-3}$. The spatial offsets $\mathbf{\Delta} = (\Delta x, \Delta y)$ are sampled from the $M$ mixture of Gaussians, described by: (1) the normalized mixture weight $\pi_j$; (2) mixture means $\bm{\mu}_j=(\mu_x, \mu_y)_j$; and (3) covariance matrices $\Sigma_j$. We further reparameterize each $\Sigma_j$ with its standard deviation $\bm{\sigma}_j =(\sigma_x, \sigma_y)_j$ and correlation coefficient $\rho_{xy,j}$.
Thus, the stroke offset distribution is 
\begin{align}
    \label{eq:stroke_offset_policy}
    p(\mathbf{\Delta}) =  \sum_{j=1}^{M}\pi_{j} \mathcal{N}(\bm{\Delta} \vert \bm{\mu}_j, \Sigma_j).
\end{align}

The RNN is implemented using a HyperLSTM~\citep{ha2013hypernetworks}; LSTM weights are generated at each timestep by a smaller recurrent ``hypernetwork'' to improve training stability. Generation is autoregressive, using $\bm{z} \in \mathbb{R}^D$, concatenated with the stroke from the previous timestep $\bm{y}_{t-1}$, to form the input to the LSTM. Stroke $\bm{y}_{t-1}$ is the ground truth supervision at train time (teacher forcing), or a sample $\bm{y'}_{t-1}$, from the mixture distribution output by the model during from timestep $t-1$.

\subsection{Training objectives}
We train the drawing model in an end-to-end fashion by jointly optimizing three losses: a pen loss $\mathcal{L}_{\text{pen}}$ for learning pen states, a stroke loss $\mathcal{L}_{\text{stroke}}$ for learning pen offsets, and our proposed pixel loss $\mathcal{L}_{\text{pixel}}$ for matching the visual similarity of the predicted and the target sketch:
\begin{align}
    \mathcal{L} = \mathcal{L}_{\text{pen}} + (1-\alpha)\mathcal{L}_{\text{stroke}} + \alpha \mathcal{L}_{\text{pixel}},
    \label{eq:total_loss}
\end{align}
where $\alpha$ is a loss weighting hyperparameter. Both $\mathcal{L}_\text{pen}$ and $\mathcal{L}_\text{stroke}$ were used in SketchRNN, while the $\mathcal{L}_\text{pixel}$ is a novel contribution to stroke-based generative models. Unlike SketchRNN, we do not impose a prior using KL divergence as we are not interested in unconditional sampling, and we found it had a negative impact on
the experiments reported below.

\paragraph{Pen loss.} The pen-states predictions $\{s_1', s_2', s_3'\}$ are optimized as a simple 3-way classification with the softmax cross-entropy loss, 
\begin{align}
    \mathcal{L}_{\text{pen}} = -\frac{1}{T}\sum_{t=1}^{T}\sum_{m=1}^3 s_{m, t}\log(s'_{m,t}).
\end{align}
\paragraph{Stroke loss.} The stroke loss maximizes the log-likelihood of the spatial offsets of each ground truth stroke $\bm{\Delta}_t$ given the mixture density distribution $p_t$ at each timestep:
\begin{align}
    \mathcal{L}_{\text{stroke}} = -\frac{1}{T}\sum_{t=1}^{T} \log p_t(\bm{\Delta}_t).
\end{align}
\paragraph{Pixel loss.}

While pixel-level reconstruction objectives are common in generative models~\citep{vae,sae,draw}, they do not exist for sketching models. However, they still represent a meaningful form of generative supervision, promoting visual similarity in the generated result. To enable this loss, we developed a novel rasterization function $f_\text{raster}$ that produces a pixel image from our stroke parameterization of sketch drawings. $f_\text{raster}$ transforms the stroke sequence $\bm{y}$ by viewing it as a set of 2D line segments ${(l_0, l_1), (l_1, l_2) \dots (l_{T-1}, l_{T})}$ where $l_t = \sum_{\tau=0}^t\bm{\Delta}_\tau$. Then, for any arbitrary canvas size we can scale the line segments, compute the distance from every pixel on the canvas to each segment and assign a pixel intensity that is inverse to the shortest distance. 

To compute the loss, we apply $f_\text{raster}$ and a Gaussian blurring filter $g_\text{blur}(\cdot)$ to both our prediction $\bm{y'}$ and ground truth $\bm{y}$ then compute the binary cross-entropy loss. The Gaussian blur is used to reduce the strictness of our pixel-wise loss.

\begin{align}
    I  = g_{\text{blur}}(&f_{\text{raster}}(\bm{y})), \quad
    I' = g_{\text{blur}}(f_{\text{raster}}(\bm{y}')) \\
    \mathcal{L}_{\text{pixel}} =& -\frac{1}{HW} \sum_{i=1}^H \sum_{j=1}^W I_{ij} \log(I'_{ij}).
\end{align}

\paragraph{Curriculum training schedule.}
We find that $\alpha$ (in Equation~\ref{eq:total_loss}) is an important hyperparameter that impacts both the learned embedding space and \model{}. A curriculum training schedule is used, increasing $\alpha$ to prioritize $\mathcal{L}_{\text{pixel}}$ relative to $\mathcal{L}_{\text{stroke}}$ as training progresses; this makes intuitive sense as a single drawing can be produced by many stroke sequences but learning to draw in a fixed manner is easier. While $\mathcal{L}_{\text{pen}}$ promotes reproducing a specific drawing sequence, $\mathcal{L}_{\text{pixel}}$ only requires that the generated drawing visually matches the image. Like a human, the model should learn to follow one drawing style (à la paint-by-numbers) before learning to draw freely.

\begin{figure}[t]
    \centering
     \begin{subfigure}{0.48\textwidth}
         \centering
         \includegraphics[width=\textwidth]{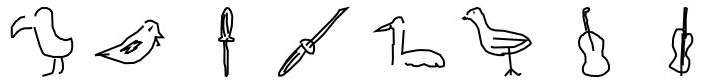}
         \caption{Quickdraw}
         \label{fig:quickdraw_samples}
     \end{subfigure}
    \begin{subfigure}{0.48\textwidth}
         \centering
         \includegraphics[width=\textwidth]{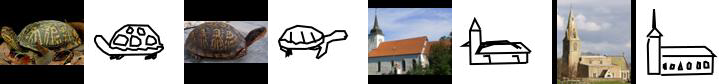}
         \caption{Sketchy}
         \label{fig:sketchy_samples}
     \end{subfigure}
     \caption{Samples of Quickdraw and Sketchy data. Sketchy examples are paired sketches and natural images.}
     \label{fig:dataset_samples}
\end{figure}

\section{Experiments}
In this section, we present our experiments on \model{} and investigate the properties of \modelembedding{}s. \model{} is trained on diverse examples of sketch--image pairs that do not include any semantic class labels. After training, we freeze the model weights and use the learned CNN encoder as the embedding function to produce \modelembedding{}s for various input images. We study the generalization of \modelembedding{}s through classification tasks involving novel examples, classes and datasets. We then examine emergent spatial and compositional properties of the representation and evaluate model generation quality. 

\subsection{Training by drawing imitation}
We train our drawing model on two different datasets that provide sketch supervision.
\begin{itemize}
\item \textbf{Quickdraw}~\citep{jongejan2016quickdraw} (Figure~\ref{fig:quickdraw_samples})
pairs sketches with a line drawing ``rendering'' of the motor program and contains 345 classes of 70,000 examples, produced by human players participating in the game ``Quick, Draw!'' 300 of 345 classes are randomly selected for training; $\bm{x}$ is rasterized to a resolution of $28 \times 28$ and stroke labels $\bm{y}$ padded up to length $T = 64$. Any drawing samples exceeding this length were discarded. Data processing procedures and class splits are in Appendix~\ref{appendix:data_processing}.

\begin{figure}[t]
        \centering
     \begin{subfigure}{0.48\textwidth}
         \centering
         \includegraphics[trim=0.05cm 0.05cm 7.5cm 12.6cm,clip,width=1\textwidth]{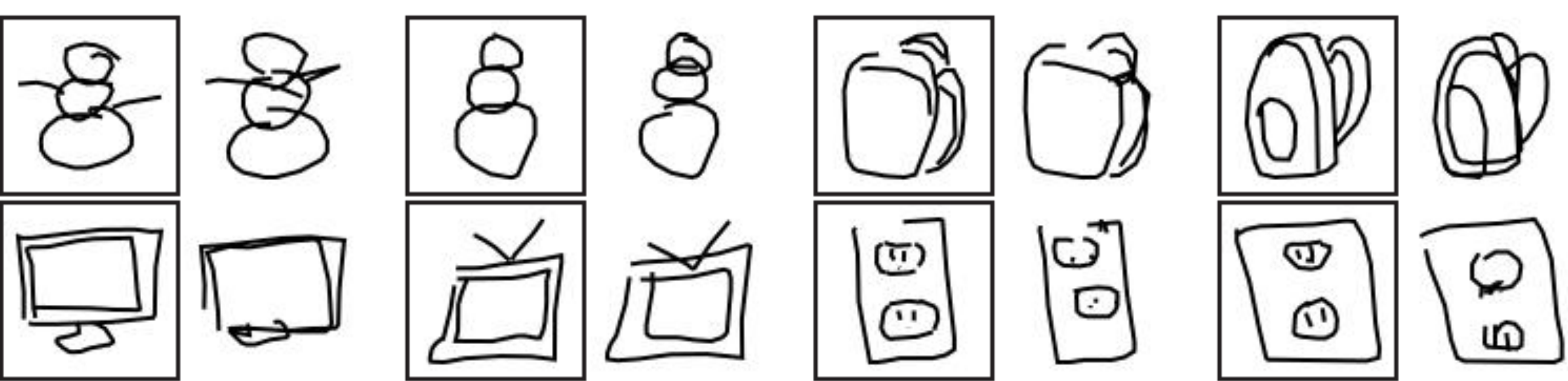}
         \caption{Quickdraw}
         \label{fig:quickdraw_sketches}
     \end{subfigure}
    \begin{subfigure}{0.48\textwidth}
         \centering
         \includegraphics[trim=0.05cm 0.05cm 7.5cm 12.6cm,clip,width=1\textwidth]{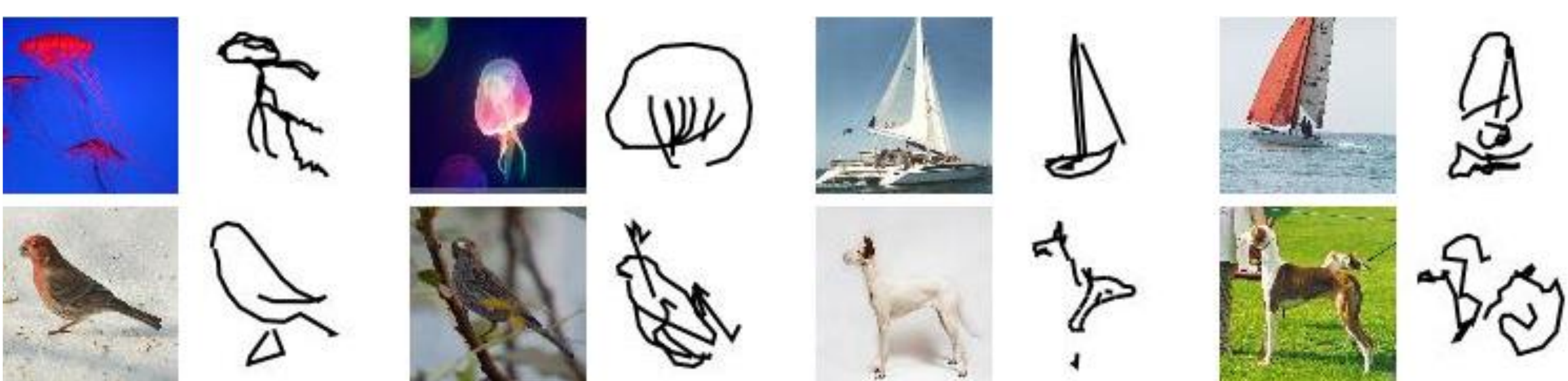}
         \caption{Sketchy}
         \label{fig:sketchy_sketches}
     \end{subfigure}
     \caption{Generated sketches of unseen examples from classes seen during training. \textbf{Left}--input; \textbf{right}--generated image}
     \label{fig:seen_sketches}
\end{figure}
\begin{table*}[t]
    \caption{Few-shot classification results on Omniglot}
    \centering
    \resizebox{0.99\textwidth}{!}{
    \begin{threeparttable}
    \begin{tabular}{@{}lllllll@{}}
\toprule
\multicolumn{3}{c}{\textbf{Omniglot}}     & \multicolumn{4}{c}{\textbf{(way, shot)}}         \\ \midrule
\textbf{Algorithm}        & \textbf{Encoder} & \textbf{Train Data} & \multicolumn{1}{c}{\textbf{(5,1)}} & \multicolumn{1}{c}{\textbf{(5,5)}} & \multicolumn{1}{c}{\textbf{(20,1)}} & \multicolumn{1}{c}{\textbf{(20,5)}}        \\ \midrule
Training from Scratch    \citep{hsu2018cactus} & N/A & Omniglot            & 52.50 $\pm$ 0.84 & 74.78 $\pm$ 0.69 & 24.91 $\pm$ 0.33 & 47.62 $\pm$ 0.44  \\
CACTUs-MAML              \citep{hsu2018cactus} & Conv4 & Omniglot            & 68.84 $\pm$ 0.80 & 87.78 $\pm$ 0.50 & 48.09 $\pm$ 0.41 & 73.36 $\pm$ 0.34  \\
CACTUs-ProtoNet          \citep{hsu2018cactus} & Conv4 & Omniglot            & 68.12 $\pm$ 0.84 & 83.58 $\pm$ 0.61 & 47.75 $\pm$ 0.43 & 66.27 $\pm$ 0.37  \\
AAL-ProtoNet        \citep{antoniou2019assume} & Conv4 & Omniglot            & 84.66 $\pm$ 0.70 & 88.41 $\pm$ 0.27 & 68.79 $\pm$ 1.03 & 74.05 $\pm$ 0.46  \\
AAL-MAML            \citep{antoniou2019assume} & Conv4 & Omniglot            & 88.40 $\pm$ 0.75 & 98.00 $\pm$ 0.32 & 70.20 $\pm$ 0.86 & 88.30 $\pm$ 1.22  \\
UMTRA              \citep{khodadadeh2019umtra} & Conv4 & Omniglot            & 83.80            & 95.43            & 74.25            & 92.12             \\ 
Random CNN                                     & Conv4 & N/A                 & 67.96 $\pm$ 0.44 & 83.85 $\pm$ 0.31 & 44.39 $\pm$ 0.23 & 60.87 $\pm$ 0.22  \\
Conv-VAE                                       & Conv4 & Omniglot            & 77.83 $\pm$ 0.41 & 92.91 $\pm$ 0.19 & 62.59 $\pm$ 0.24 & 84.01 $\pm$ 0.15  \\
Conv-VAE                                       & Conv4 & Quickdraw           & 81.49 $\pm$ 0.39 & 94.09 $\pm$ 0.17 & 66.24 $\pm$ 0.23 & 86.02 $\pm$ 0.14  \\
\hline
Contrastive                                    & Conv4 & Omniglot*           & 77.69 $\pm$ 0.40 & 92.62 $\pm$ 0.20 & 62.99 $\pm$ 0.25 & 83.70 $\pm$ 0.16 \\
\model{} \textit{(Ours)}                       & Conv4 & Omniglot*           & \textbf{94.88} $\pm$ 0.22 & \textbf{99.01} $\pm$ 0.08 & \textbf{86.18} $\pm$ 0.18 & \textbf{96.69} $\pm$ 0.07  \\
Contrastive                                    & Conv4 & Quickdraw*          & 83.26 $\pm$ 0.40 & 94.16 $\pm$ 0.21 & 73.01 $\pm$ 0.25 & 86.66 $\pm$ 0.17  \\
\model{} \textit{(Ours)}                       & Conv4 & Quickdraw*          & \textbf{96.96} $\pm$ 0.17 & \textbf{99.50} $\pm$ 0.06 & \textbf{91.67} $\pm$ 0.14 & \textbf{98.30} $\pm$ 0.05  \\ \midrule
MAML \textit{(Supervised)} \citep{finn2017maml}                 & Conv4 & Omniglot            & 94.46 $\pm$ 0.35 & 98.83 $\pm$ 0.12 & 84.60 $\pm$ 0.32 & 96.29 $\pm$ 0.13  \\
ProtoNet \textit{(Supervised)} \citep{snell2017prototypical}    & Conv4 & Omniglot            & 98.35 $\pm$ 0.22 & 99.58 $\pm$ 0.09 & 95.31 $\pm$ 0.18 & 98.81 $\pm$ 0.07  \\ \bottomrule
    \end{tabular}
    \begin{tablenotes}\footnotesize
    \item[*] Sequential sketch supervision used for training
    \end{tablenotes}
    \end{threeparttable}}
    \label{tab:omniglot_results}
\end{table*}
\item \textbf{Sketchy}~\citep{Sangkloy2016sketchy} (Figure~\ref{fig:sketchy_samples}) is a more challenging collection of (photorealistic) natural image--sketch pairs and contains 125 classes from ImageNet~\citep{imagenet}, selected for ``sketchability''. Each class has 100 natural images paired with up to 20 loosely aligned sketches for a total of 75,471 image--sketch pairs. Images are resized to $84 \times 84$ and padded to increase spatial agreement; sketch sequences are set to a max length $T=100$. Classes 
that overlap with 
the test set of mini-ImageNet~\citep{ravi2016optimization} are removed from our training set, to faithfully evaluate few-shot classification performance.
\end{itemize}

Data samples are presented in Figure~\ref{fig:dataset_samples}; for Quickdraw, the input image $\bm{x}$ and the rendered sketch $\bm{y}$ are the same.

We train a single model on Quickdraw using a 4-layer CNN (Conv4) encoder~\cite{vinyals2016matching} and another on the Sketchy dataset with a ResNet-12~\citep{oreshkin2018tadam} encoder architecture. 

\paragraph{Baselines.}
We consider the following baselines to compare with \model{}. 
\begin{itemize}
\item \textbf{Contrastive} is similar to the search embedding of \citet{sketchformer}; a metric learning baseline that matches CNN image embeddings with corresponding RNN sketch embeddings. Our baseline is trained using the InfoNCE loss~\citep{Oord2018infonce}.

\item \textbf{Conv-VAE}~\citep{vae} performs pixel-level representation learning without motor program information. 

\item \textbf{Pix2Pix}~\citep{pix2pix} is a generative adversarial approach that performs image to sketch domain transfer but is supervised by sketch images and not the sequential motor program.
\end{itemize}

Note that Contrastive is an important comparison for \model{} as it also uses the motor-program sequence when training on sketch-image pairs.



\paragraph{Implementation details.}
\model{} is trained for 300k iterations with batch size of 256 for Quickdraw and 64 for Sketchy due to memory constraints. Initial learning rate is 1e-3 decaying by $0.85$ every 15k steps. We use the Adam~\citep{adam} optimizer and clip gradient values to $1.0$. Latent space $\text{dim}(z) = 256$, RNN output size is $1024$, and hypernetwork embedding is 64. Mixture count is $M = 30$ and Gaussian blur from $\mathcal{L}_\text{pixel}$ uses $\sigma=2.0$.

Conv4 encoder is identical to ~\citet{vinyals2016matching} and the ResNet-12 encoder uses $4$ blocks of 64-128-256-512 filters with ReLU activations. $\alpha$ is set to $0$ and increases by $0.05$ every 10k training steps with an empirically obtained cap at $\alpha_\text{max}=0.50$ for Quickdraw and $\alpha_\text{max}=0.75$ for Sketchy. 
See Appendix~\ref{appendix:implementation_details} for additional details.


\subsection{Few-Shot Classification using \modelembedding{}s}
\model{} transforms images to strokes, the learned, shape-biased representations could be useful for explaining a novel concept.
In this section, we evaluate the ability of learning novel concepts from unseen datasets using few-shot classification benchmarks on Omniglot~\citep{Lake2015bpl} and mini-ImageNet~\citep{vinyals2016matching}. In few-shot classification,
models learn a set of novel classes from only a few examples.
We perform few-shot learning on standard $N$-way, $K$-shot episodes by training a simple linear classifier on top of \modelembedding{}s.

\begin{table*}
    \caption{Few-shot classification results on mini-ImageNet}
    \centering
    \resizebox{0.99\textwidth}{!}{
    \begin{threeparttable}
    \begin{tabular}{@{}lllllll@{}}
\toprule
\multicolumn{3}{c}{\textbf{mini-ImageNet}}& \multicolumn{4}{c}{\textbf{(way, shot)}}         \\ \midrule
\textbf{Algorithm}       & \textbf{Backbone} & \textbf{Train Data} & \multicolumn{1}{c}{\textbf{(5,1)}} & \multicolumn{1}{c}{\textbf{(5,5)}} & \multicolumn{1}{c}{\textbf{(5,20)}} & \multicolumn{1}{c}{\textbf{(5,50)}}        \\ \midrule
Training from Scratch   \citep{hsu2018cactus} & N/A & mini-ImageNet       & 27.59 $\pm$ 0.59 & 38.48 $\pm$ 0.66 & 51.53 $\pm$ 0.72 & 59.63 $\pm$ 0.74  \\ 
CACTUs-MAML              \citep{hsu2018cactus} & Conv4 & mini-ImageNet       & 39.90 $\pm$ 0.74 & 53.97 $\pm$ 0.70 & 63.84 $\pm$ 0.70 & 69.64 $\pm$ 0.63  \\
CACTUs-ProtoNet          \citep{hsu2018cactus} & Conv4 & mini-ImageNet       & 39.18 $\pm$ 0.71 & 53.36 $\pm$ 0.70 & 61.54 $\pm$ 0.68 & 63.55 $\pm$ 0.64  \\
AAL-ProtoNet       \citep{antoniou2019assume} & Conv4 & mini-ImageNet       & 37.67 $\pm$ 0.39 & 40.29 $\pm$ 0.68 &\multicolumn{1}{c}{-}&\multicolumn{1}{c}{-} \\
AAL-MAML           \citep{antoniou2019assume} & Conv4 & mini-ImageNet       & 34.57 $\pm$ 0.74 & 49.18 $\pm$ 0.47 &\multicolumn{1}{c}{-}&\multicolumn{1}{c}{-} \\
UMTRA             \citep{khodadadeh2019umtra} & Conv4 & mini-ImageNet       & 39.93   & 50.73            & 61.11            & 67.15             \\
Random CNN                                   & Conv4 & N/A       & 26.85 $\pm$ 0.31 & 33.37 $\pm$ 0.32 & 38.51 $\pm$ 0.28 & 41.41 $\pm$ 0.28  \\ 
Conv-VAE                                     & Conv4 & mini-ImageNet       & 23.30 $\pm$ 0.21 & 26.22 $\pm$ 0.20 & 29.93 $\pm$ 0.21 & 32.57 $\pm$ 0.20  \\
Conv-VAE                                     & Conv4 & Sketchy       & 23.27 $\pm$ 0.18 & 26.28 $\pm$ 0.19 & 30.41 $\pm$ 0.19 & 33.97 $\pm$ 0.19  \\ 
Random CNN                                   & ResNet12 & N/A                & 28.59 $\pm$ 0.34 & 35.91 $\pm$ 0.34 & 41.31 $\pm$ 0.33 & 44.07 $\pm$ 0.31  \\ 
Conv-VAE                                     & ResNet12 & mini-ImageNet      & 23.82 $\pm$ 0.23 & 28.16 $\pm$ 0.25 & 33.64 $\pm$ 0.27 & 37.81 $\pm$ 0.27  \\
Conv-VAE                                     & ResNet12 & Sketchy            & 24.61 $\pm$ 0.23 & 28.85 $\pm$ 0.23 & 35.72 $\pm$ 0.27 & 40.44 $\pm$ 0.28  \\
\hline
Contrastive                  & ResNet12 & Sketchy*           & 30.56 $\pm$ 0.33 & 39.06 $\pm$ 0.33 & 45.17 $\pm$ 0.33 & 47.84 $\pm$ 0.32  \\ 
\model{} \textit{(ours)}              & Conv4 & Sketchy*                & 38.61 $\pm$ 0.42 & 53.82 $\pm$ 0.41 & 63.34 $\pm$ 0.35 & 67.22 $\pm$ 0.32 \\
\model{} \textit{(ours)}              & ResNet12 & Sketchy*             & \textbf{40.39} $\pm$ 0.44 & \textbf{57.15} $\pm$ 0.38 & \textbf{67.60} $\pm$ 0.33 & \textbf{71.99} $\pm$ 0.3 \\\midrule
MAML \textit{(supervised)} \citep{finn2017maml}                 & Conv4 & mini-ImageNet       & 46.81 $\pm$ 0.77 & 62.13 $\pm$ 0.72 & 71.03 $\pm$ 0.69 & 75.54 $\pm$ 0.62  \\
ProtoNet \textit{(supervised)} \citep{snell2017prototypical}    & Conv4 & mini-ImageNet       & 46.56 $\pm$ 0.76 & 62.29 $\pm$ 0.71 & 70.05 $\pm$ 0.65 & 72.04 $\pm$ 0.60  \\ \bottomrule
    \end{tabular}
    \begin{tablenotes}\footnotesize
    \item[*] Sequential sketch supervision used for training
    \end{tablenotes}
    \end{threeparttable}}
    \label{tab:mini-imagenet_results}
    \vspace{-0.15in}
\end{table*}
\begin{table}[t]
\begin{center}
    \caption{Effect of $\alpha_\text{max}$ on few-shot classification accuracy}
    \label{tab:alpha_sweep}
    \resizebox{0.48\textwidth}{!}{\begin{tabular}{c|cccccc}
        $\alpha_\text{max}$ & 0.00 & 0.25 & 0.50 & 0.75 & 0.95 & 1.00     \\ \midrule
        Omniglot(20,1) & 87.17 & 87.82 & \textbf{91.67} & 90.59 & 89.77 & 87.63 \\
        mini-ImageNet(5,1) & 38.00 & 38.75 & 38.11 & \textbf{39.31} & 38.53 & 37.78 \\
    \end{tabular}}
\end{center}
\end{table}

Typically, the training data of few-shot classification is fully labelled, and the standard approaches learn by
utilizing the labelled training data before evaluation on novel test classes~\cite{vinyals2016matching,finn2017maml,snell2017prototypical}.
Unlike these methods, \model{} does not use class labels during training. Therefore, we compare our model to unsupervised few-shot learning methods \textbf{CACTUs}~\citep{hsu2018cactus}, \textbf{AAL}~\citep{antoniou2019assume} and \textbf{UMTRA}~\citep{khodadadeh2019umtra}. CACTUs is a clustering-based method while AAL and UMTRA use data augmentation to approximate supervision for meta-learning~\citep{finn2017maml}. We also compare to our baselines that use this sketch information:
both \model{} and \textbf{Contrastive} use motor-program sequence supervision, and \textbf{Pix2Pix}~\citep{pix2pix} requires natural and sketch image pairings.
In addition to these, we provide supervised few-shot learning results using \textbf{MAML}~\citep{finn2017maml} and \textbf{ProtoNet}~\citep{snell2017prototypical} as references.

\paragraph{Omniglot results.} The results on Omniglot~\citep{Lake2015bpl} using the split from~\citet{vinyals2016matching} are reported in Table~\ref{tab:omniglot_results}. \model{} obtains the highest classification accuracy when training on the Omniglot dataset. The Conv-VAE and as well as the Contrastive model are outperformed by existing unsupervised methods but not by a huge margin.\footnote{We do not include the Pix2Pix baseline here as the input and output images are the same.} When training on the Quickdraw dataset \model{} sees a substantial accuracy increase and exceeds the classification accuracy of the supervised MAML approach. While our model has arguably more supervision information than the unsupervised methods, our performance gains relative to the Contrastive baseline shows that this does not fully explain the results. Furthermore, our method transfers well from Quickdraw to Omniglot without ever seeing a single Omniglot character.
\begin{table}[t]
    \centering
    \caption{Classification accuracy of novel examples and classes}
    \label{tab:intradataset_classification}
    \begin{subtable}[t]{0.48\textwidth}
        \centering
        \resizebox{0.83\textwidth}{!}{\begin{tabular}{c|c|c}
             & 300-way & 45-way  \\ 
            & Training Classes & Unseen Classes \\ \midrule
            Random CNN & 0.85 & 16.42 \\
            Conv-VAE & 18.70 & 53.06 \\
            Contrastive 
            & 41.58 & 70.72 \\
            \model{} & \textbf{42.80} &  \textbf{75.68} \\
        \end{tabular}}
        \caption{Quickdraw}
        \label{tab:classification_quickdraw}
    \end{subtable}\\
    \begin{subtable}[t]{0.48\textwidth}
        \centering
        \resizebox{0.83\textwidth}{!}{\begin{tabular}{c|c|c}
            Embedding model & ILSVRC Top-1 & ILSVRC Top-5  \\ \midrule
            Random CNN & 1.58 & 4.78 \\
            Conv-VAE & 1.13 & 3.78 \\
            Pix2Pix & 1.23 & 4.29\\
            Contrastive & 3.95 & 10.91 \\
            \model{} & \textbf{6.15} &  \textbf{16.20} \\
        \end{tabular}}
        \caption{Sketchy. 
        }
        \label{tab:classification_sketchy}
    \end{subtable}
\end{table}

\paragraph{mini-ImageNet results.} The results on mini-ImageNet~\citep{vinyals2016matching} using the split from~\cite{ravi2016optimization} are reported in Table~\ref{tab:mini-imagenet_results}. \model{} outperforms existing unsupervised few-shot classification approaches. 
We report results using both Conv4 and ResNet12 backbones; the latter allows more learning capacity for the drawing imitation task, and consistently achieves better performance.
Unlike on the Omniglot benchmark, Contrastive and Conv-VAE perform poorly compared to existing methods, whereas \model{} scales well to natural images and again outperforms other unsupervised few-shot learning methods, and even matches the performance of a supervised ProtoNet on 5-way 50-shot (71.99 vs. 72.04).
This suggests that forcing the model to generate sketches yields more informative representations.
{\bf Effect of pixel-loss weighting.} We ablate pixel loss coefficient $\alpha_\text{max}$ to quantify its impact on the observed representation, using the Omniglot task (Table \ref{tab:alpha_sweep}). There is a substantial improvement in few-shot classification when $\alpha_\text{max}$ is non-zero. $\alpha_\text{max}$= 0.50 achieves the best results for Quickdraw, while it trends downwards when $\alpha_\text{max}$ approaches to 1.0. mini-ImageNet performs best at $\alpha_\text{max}=0.75$ Over-emphasizing the pixel-loss while using teacher forcing causes the model to create sketches by using many strokes, and does not generalize to true autoregressive generation.

\subsection{Intra-Dataset Classification}
\label{sec:intra-dataset_classification}

\begin{figure}[t]
    \centering
    \includegraphics[trim=6.1cm 0.2cm 7cm 0cm,clip,width=0.48\textwidth]{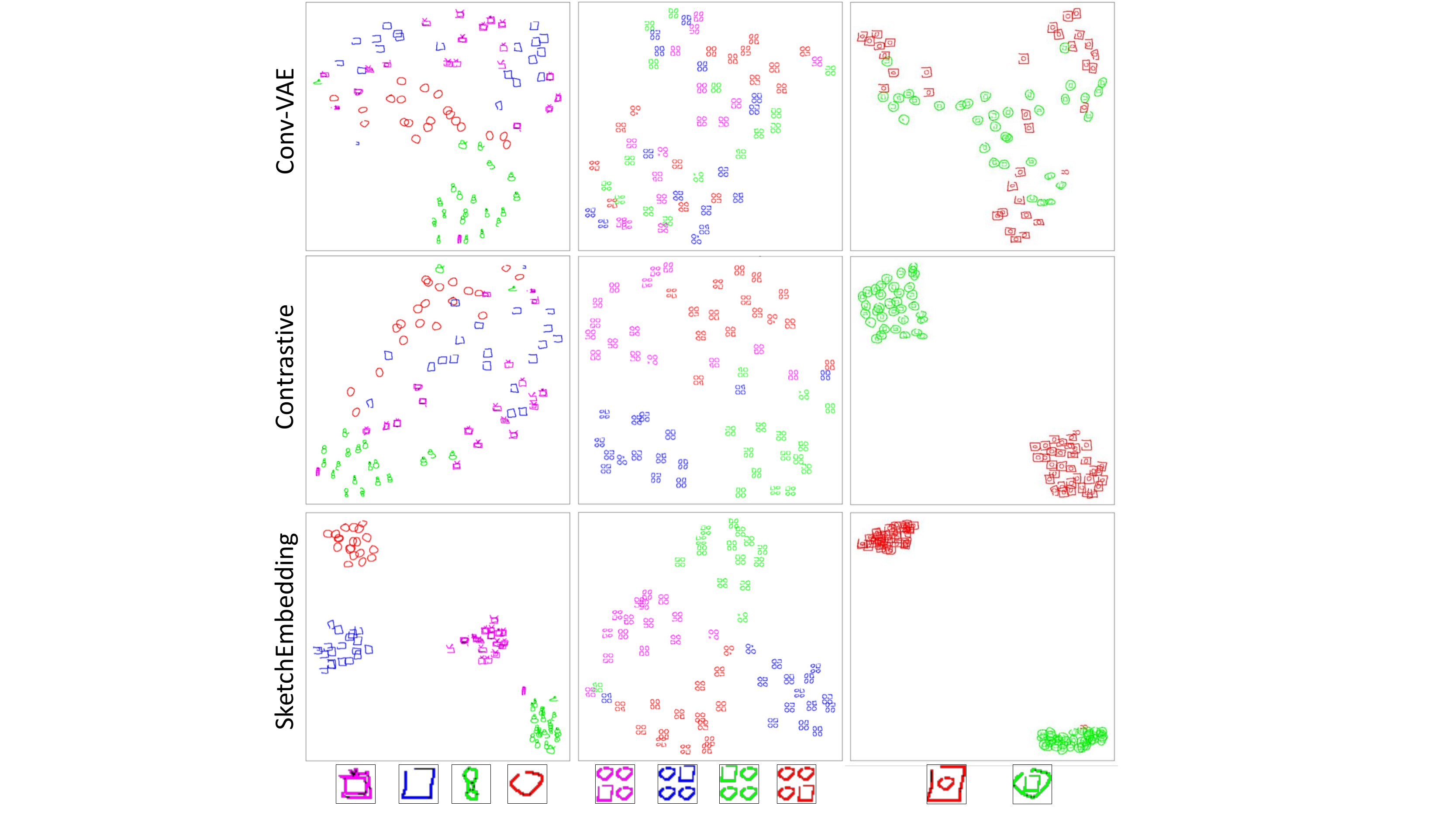}
    \caption{Embedding clustering of images with different component arrangements. \textbf{Left}--numerosity; \textbf{middle}--placement; \textbf{right}--containment}
    \label{fig:compositionality_component_arrangements}
\end{figure}
\begin{figure}[t]
    \centering
    \includegraphics[trim=6.1cm 0.4cm 7cm 0cm,clip,width=0.48\textwidth]{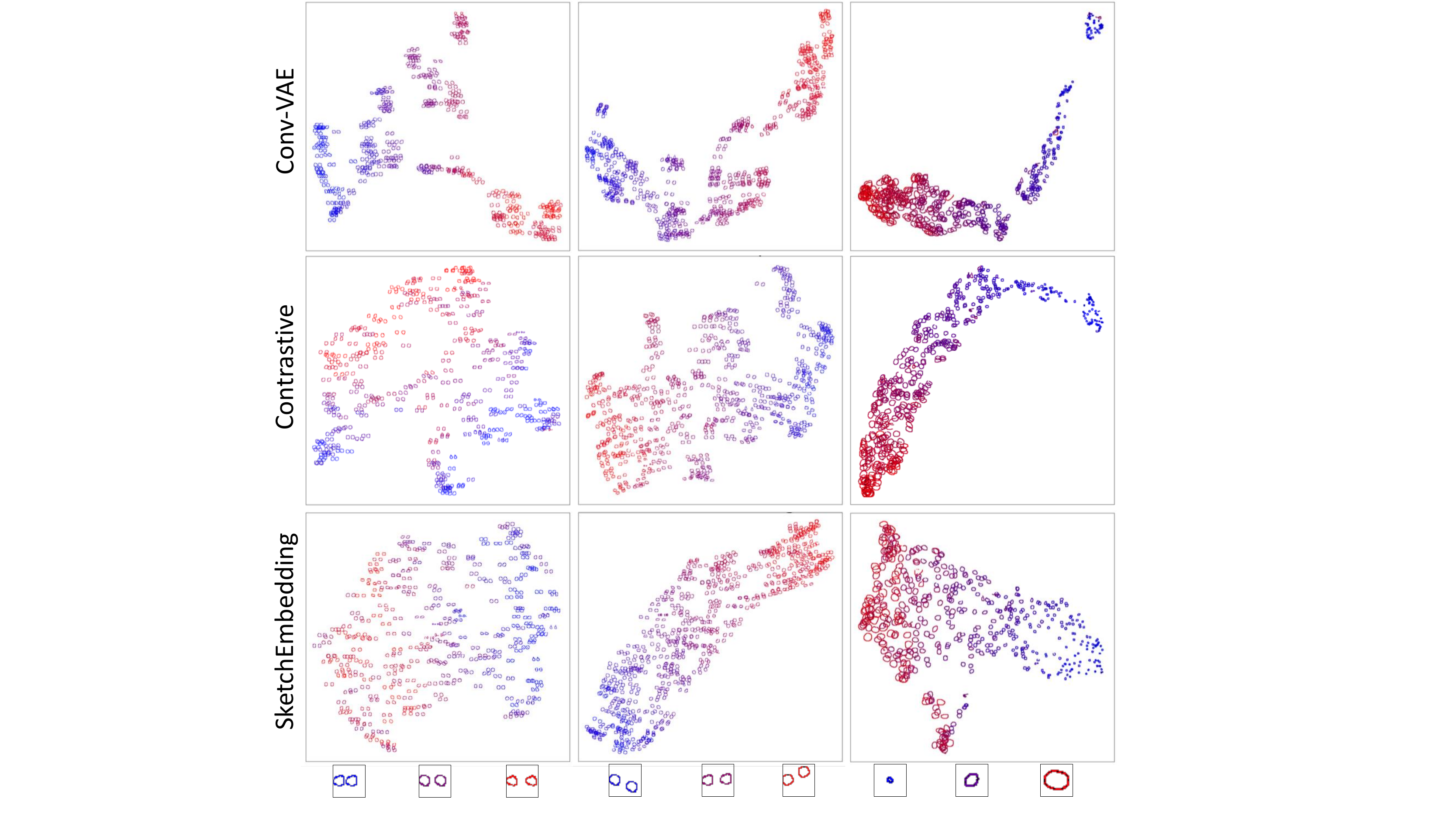}
    \caption{Recovering spatial variables embedded within image components. \textbf{Left}--distance; \textbf{middle}--angle; \textbf{right}--size}
    \label{fig:compositionality_spatial_relationships}
\end{figure}

While few-shot classification demonstrates a strong form of generalization to novel classes, and in \model{}'s case entirely new datasets, we also investigate the useful information learned from the same datasets used in training. Here we study a conventional classification problem:
we train a 
single layer linear classifier on top of input \modelembedding{}s of images drawn from the training dataset. We report accuracy on a validation set of novel images from the same classes, or new classes from the same training dataset.

\paragraph{Quickdraw results.} The training data consists of 256 labelled examples for each of the 300 training classes. New example generalization is evaluated in 300-way classification on unseen examples of training classes. Novel class generalization is evaluated on 45-way classification of unseen Quickdraw classes. The results are presented in Table~\ref{tab:classification_quickdraw}. \model{} obtains the best classification performance. The Contrastive method also performs well, demonstrating the informativeness of sketch supervision. Note that while Contrastive performs well on training classes, it performs worse on unseen classes. The few-shot benchmarks in Tables~\ref{tab:omniglot_results},~\ref{tab:mini-imagenet_results} suggest our generative objective is more suitable for novel class generalization. Unlike in the few-shot tasks, a Random CNN performs very poorly likely because the linear classification head lacks the capacity to discriminate the random embeddings.

\paragraph{Sketchy results.} 
Since there are not enough examples or classes to test unseen classes within Sketchy, we evaluate model generalization on 1000-way classification of ImageNet-1K (ILSVRC2012), and the validation accuracy is presented in Table~\ref{tab:classification_sketchy}.
It is  important to note that all the methods shown here only have access to a maximum of 125 Sketchy classes during training, resized down to 84$\times$84, with a max of 100 unique photos per class, and thus they are not directly comparable to current state-of-the-art methods trained on ImageNet. \model{} once again obtains the best performance,
not only relative to the image-based baselines, Random CNN, Conv-VAE and Pix2Pix, but also to the Contrastive learning model, which like \model{} utilizes the sketch information during training. 
While Contrastive is competitive in Quickdraw classification, it does not maintain this performance on more difficult tasks with natural images, much like in the few-shot natural image setting. Unlike in Quickdraw classification where pretraining is effective, all 3 pixel-based methods perform similarly poorly.

\subsection{Emergent properties of \modelembedding{}s}
\label{sec:compositionality}
Here we probe properties of the image representations formed by \model{} and the baseline models. We construct a set of experiments to showcase the spatial and component-level visual understanding and conceptual composition in the embedding space.

\paragraph{Arrangement of image components.}
To test component-level awareness, we construct image examples containing different arrangements of multiple objects in image space.  We then embed these examples and project into 2D space using UMAP~\citep{mcinnes2018umap} to visualize their organization. The leftmost panel of Figure \ref{fig:compositionality_component_arrangements} exhibits a numerosity relation with
Quickdraw classes containing duplicated components; snowmen with circles and televisions with squares. The next two panels of Figure \ref{fig:compositionality_component_arrangements} contain 
examples with a placement and containment relation. \modelembedding{} representations are the most distinguishable and are easily separable. The pixel-based Conv-VAE is the least distinguishable, while the Contrastive model 
performs well in the containment case but poorly in the other two. 
As these image components are drawn contiguous through time and separated by lifted pen states, \model{} learns to group the input pixels together as abstract elements to be drawn together.

\begin{figure}[t]
    \centering
    \includegraphics[trim=0cm 0cm 21.2cm 11.5cm,clip,width=0.48\textwidth]{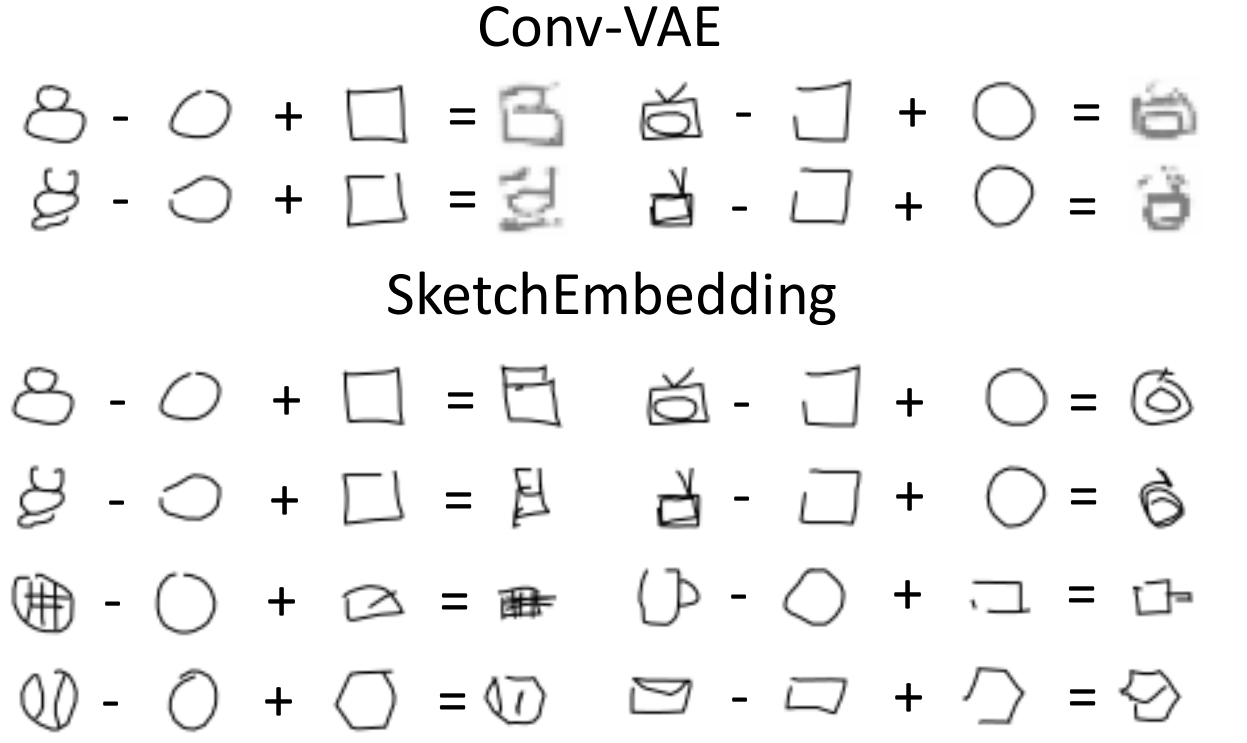}
    \caption{Conceptual composition of image representations. Several sketches are shown for the two models following algebraic operations on their embedding vectors.}
    \label{fig:compositionality_conceptual_composition}
\end{figure}

\paragraph{Recovering spatial relationships.}
We examine how the underlying variables of distance, angle or size are captured by the studied embedding functions. We construct and embed examples changing each of the variables of interest. The embeddings are again projected into 2D by the UMAP~\citep{mcinnes2018umap} algorithm in Figure \ref{fig:compositionality_spatial_relationships}. After projection, \model{} recovers the variable of interest as an approximately linear manifold in 2D space; the Contrastive embedding produces similar results, while the pixel-based Conv-VAE is more clustered and non-linear. This shows that relating images to sketch motor programs encourages the system to learn the spatial relationships between components, since it needs to produce the $\Delta \bm{x} $ and $ \Delta \bm{y}$ values to satisfy the training objective.

\paragraph{Conceptual composition.}
Finally, we explore the use of \modelembedding{}s for composing embedded concepts. In natural language literature, vector algebra such as ``king'' - ``man'' + ``woman'' = ``queen''~\citep{mikolov2013cbow} shows linear compositionality in the concept space of word embedding. It has also been demonstrated in human face images and vector graphics~\citep{bojanowski2017optimizing,shen2020interpreting,carlier2020deepsvg}. Here we try to explore such concept compositionality property in sketch image understanding as well.
We embed examples of simple shapes such as a square or circle as well as more complex examples like a snowman or mail envelope and perform arithmetic in the latent space. Surprisingly, upon decoding the \modelembedding{} vectors we recover intuitive sketch generations. For example, if we subtract the embedding of a circle from snowman and add a square, then the resultant vector gets decoded into an image of a stack of boxes. We present examples in Figure~\ref{fig:compositionality_conceptual_composition}. By contrast, the Conv-VAE does not produce sensible decodings on this task.

\begin{figure}[t]
    \centering
    \includegraphics[trim=0.05cm 0.05cm 7cm 12.6cm,clip,width=0.48\textwidth]{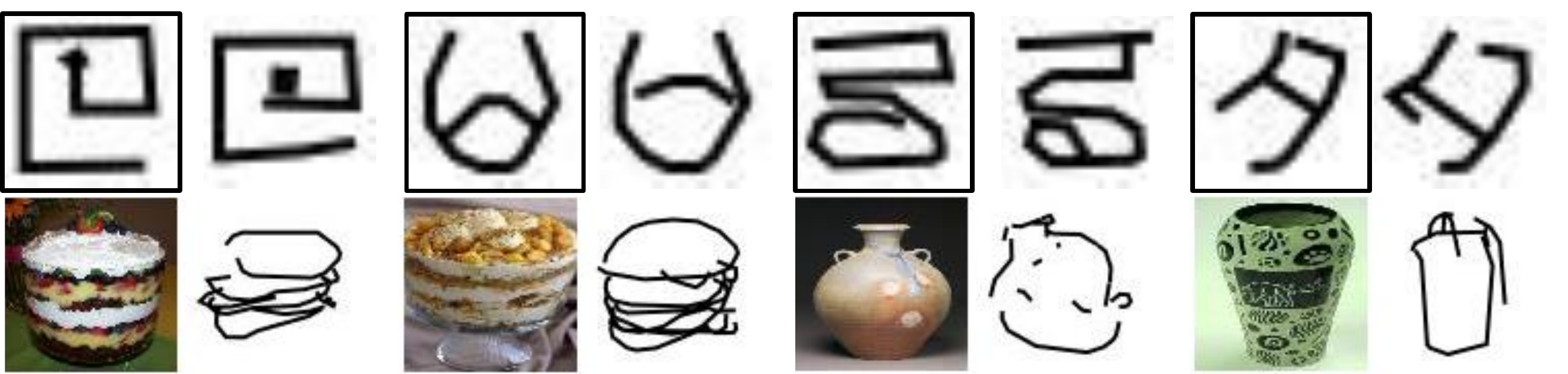}
    \caption{Generated sketches of images from datasets unseen during training. \textbf{Left}--input; \textbf{right}--generated image}
    \label{fig:few_shot_samples}
\end{figure}

\begin{table}[t]
    \centering
    \begin{tabular}{ccc}
            \toprule
             & Seen  & Unseen \\ \midrule
            Original Data         & 97.66         & 96.09         \\ \midrule
            Conv-VAE              & 76.28 $\pm$ 0.93 & 75.07 $\pm$ 0.84 \\
            \model{}       & {\bf 81.44} $\pm$ 0.95 & {\bf 77.94} $\pm$ 1.07 \\ \bottomrule
            \end{tabular}
    \caption{Classification accuracy for generated sketch images.}
    \label{tab:generative_scores}
\end{table}

\begin{figure*}[t]
    \centering
    \includegraphics[trim=0cm 0.22cm 0cm 0.2cm,clip,width=0.75\textwidth]{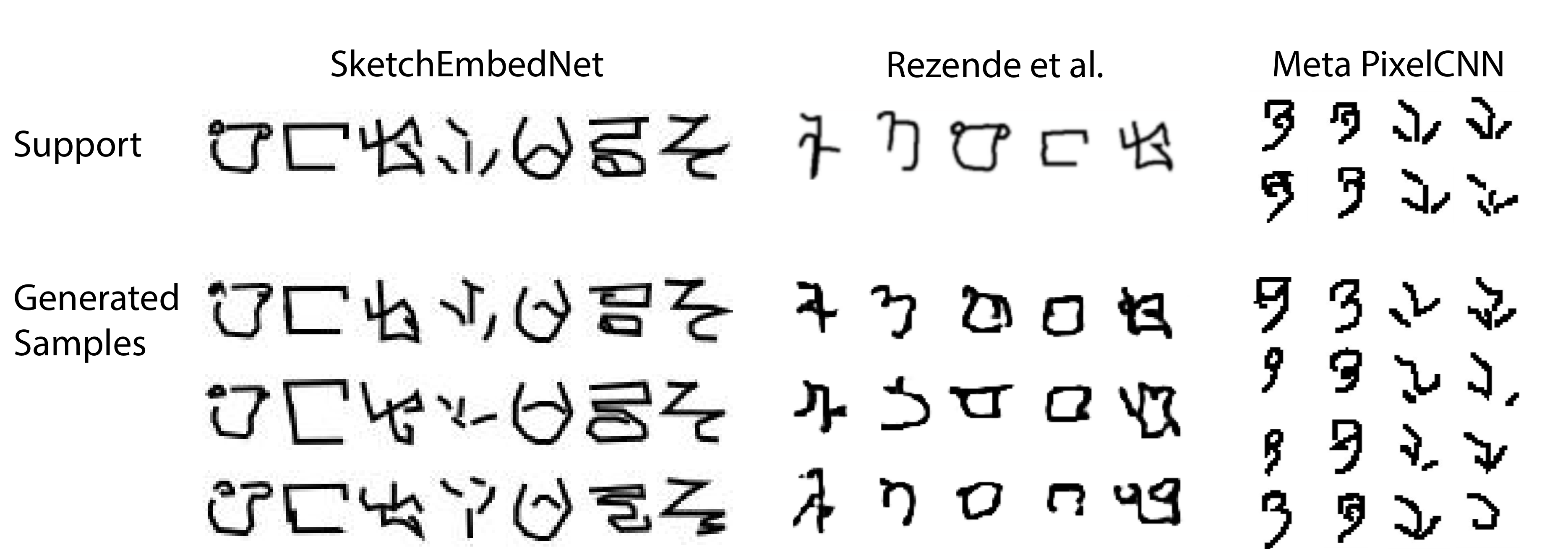}
    \caption{One-shot Omniglot generation compared to ~\citet{oneshotgen, fewshotgen}.}
    \label{fig:otherpapergenerations}
\end{figure*}
\subsection{Evaluating generation quality}
\label{sec:generation_quality}
Another method to evaluate our learned image representations is through the sketches generated based on these representations; a good representation should produce a recognizable image. Figures \ref{fig:seen_sketches} and \ref{fig:few_shot_samples} show that \model{} can generate reasonable sketches of training classes as well as unseen data domains. When drawing natural images, it sketches the general shape of the subject rather than replicating specific details.  

\paragraph{Classifying generated examples.}
Quantative assessment of generated images is often challenging and per-pixel metrics like in ~\cite{fewshotgen, oneshotgen} may penalize generative variation that still preserves meaning. We train ResNet classifiers
for an Inception Score~\citep{inception} inspired metric. One classifier is trained on 45  (``seen'') Quickdraw training classes and the other on 45 held out (``unseen'') classes that were not encountered during model training. Samples generated by a sketching model are rendered, then classified;  we report each classifier's accuracy on these examples compared to its training accuracy in Table \ref{tab:generative_scores}. 
\model{} produces more recognizable sketches than a Conv-VAE model when generating examples of both seen and unseen object classes.

\paragraph{Qualitative comparison of generations.}
In addition to the Inception-score~\citep{inception} inspired metric, we also qualitatively assess the generations of \model{} on unseen datasets. One-shot generations are sampled from Omniglot~\citep{Lake2015bpl} and are visually compared with other few- and one-shot generation methods~\citep{oneshotgen, fewshotgen} (Figure \ref{fig:otherpapergenerations}). 

None of the models have seen any examples from the character class or parent alphabet. Furthermore, \model{} was not trained on any Omniglot data. Visually, our generated images better resemble the support examples and have generative variance that better preserves class semantics. Generations in pixel space may disrupt strokes and alter the character to human perception. This is especially true for written characters as they are frequently defined by a specific set of strokes instead of blurry clusters of pixels.

\paragraph{Discussion.}
While having a generative objective is useful for representation learning (we see that \model{} outperform our Contrastive representations), it is insufficient to guarantee an informative embedding for other tasks. The Conv-VAE generations perform slightly worse on the recognizability task in Table~\ref{tab:generative_scores}, 
while being significantly worse in our previous classification tasks in Tables~\ref{tab:omniglot_results}, \ref{tab:mini-imagenet_results} and~\ref{tab:intradataset_classification}.

This suggests that the output domain has an impact on the learned representation. The increased componential and spatial awareness from generating sketches (as in Section~\ref{sec:compositionality}) makes \modelembedding{}s better for downstream classification tasks by better capturing the visual shape in images.

\section{Conclusion}
Learning to draw is not only an artistic pursuit but drives a distillation of real-world visual concepts. 
In this paper, we present a model that learns representation of images 
which capture salient features, by producing
sketches of image inputs.
While sketch data may be challenging to source, we show that \model{} can generalize to image domains beyond the training data.
Finally, \model{} achieves competitive performance on few-shot learning of novel classes, and represents compositional properties,
suggesting that learning to draw can be a promising avenue for learning general visual representations.

\paragraph{Acknowledgments}
We thank Jake Snell, James Lucas and Robert Adragna for their helpful feedback on earlier drafts of
the manuscript. 
Resources used in preparing this research were provided, in part, by the
Province of Ontario, the Government of Canada through CIFAR, and companies
sponsoring the Vector Institute (\url{www.vectorinstitute.ai/\#partners}).
This project is supported by NSERC and the Intelligence Advanced Research Projects
Activity (IARPA) via Department of Interior/Interior Business Center (DoI/IBC) contract number
D16PC00003. The U.S. Government is authorized to reproduce and distribute reprints for Governmental
purposes notwithstanding any copyright annotation thereon. Disclaimer: The views and conclusions
contained herein are those of the authors and should not be interpreted as necessarily representing
the official policies or endorsements, either expressed or implied, of IARPA, DoI/IBC, or the U.S.
Government.

\clearpage

\bibliography{main}
\bibliographystyle{icml2021}

\clearpage

\appendix
\section{Rasterization}
\label{appendix:rasterization}
The key enabler of our novel pixel loss for sketch drawings is our differentiable rasterization function $f_\text{raster}$. Sequence based loss functions such as $\mathcal{L}_\text{stroke}$ are sensitive to the order of points while in reality, drawings are sequence invariant. Visually, a square is a square whether it is drawn clockwise or counterclockwise. 

One purpose of the sketch representation is to lower the complexity of the data space and decode in a more visually intuitive manner. While it is a necessary departure point, the sequential generation of drawings is not key to our visual representation and we would like \model{} to be agnostic to any specific sequence needed to draw the sketch that is representative of the image input.

To facilitate this, we develop our rasterization function $f_\text{raster}$ which renders an input sequence of strokes as a pixel image. However, during training, the RNN outputs a mixture of Gaussians at each timestep. To convert this to a stroke sequence, we sample from these Gaussians; this can be repeated to reduce the variance of the pixel loss.
We then scale our predicted and ground truth sequences by the properties of the latter before rasterization.

\paragraph{Stroke sampling.} At the end of sequence generation we have $N_s \times (6M + 3)$ parameters, $6$ Gaussian mixture parameters, 3 pen states, $N_s$ times, one for each stroke. To obtain the actual drawing we sample from the mixture of Gaussians:
\begin{align}
     \begin{bmatrix} \Delta x_t \\  \Delta y_t \end{bmatrix} &=  \begin{bmatrix}
        \mu_{x,t} \\ \mu_{y,t} 
    \end{bmatrix}
     + 
     \begin{bmatrix}
        \sigma_{x,t} & 0 \\ 
        \rho_{xy,t} \sigma_{y,t} & \sigma_{y,t} \sqrt{1 - \rho_{xy,t}^2}
     \end{bmatrix} \bm{\epsilon} \\ \bm{\epsilon} &\sim \mathcal{N}(\bm{0}, \bm{1}_2).
\end{align}
After sampling we compute the cumulative sum of every stroke over the time so that we obtain an absolute position at each timestep:
\begin{align}
    \begin{bmatrix} x_t \\ y_t \end{bmatrix} &= 
    \sum_{\tau = 0}^{T}\begin{bmatrix} \Delta x_\tau \\  \Delta y_\tau \end{bmatrix}. \label{eq:abs_pos}
\end{align}
\begin{align}
    \bm{y}_{t,abs} = (x_t, y_t, s_1, s_2, s_3).
\end{align}

\paragraph{Sketch scaling.}
Each sketch generated by our model begins at (0,0) and the variance of all strokes in the training set is normalized to $1$. On a fixed canvas the image is both very small and localized to the top left corner. We remedy this by computing a scale $\lambda$ and shift $x_{\text{shift}}, y_{\text{shift}}$ using labels $\bm{y}$ and apply them to both the prediction $\bm{y}'$ as well as the ground truth $\bm{y}$. These parameters are computed as:
\begin{align}
    \lambda = \text{min} \left\{\frac{W}{x_\text{max} - x_\text{min}}, \frac{H}{y_\text{max} - y_\text{min}} \right\},
\end{align}
\begin{align}
    x_\text{shift} = \frac{x_\text{max} + x_\text{min}}{2}\lambda, \ \
    y_\text{shift} = \frac{y_\text{max} + y_\text{min}}{2}\lambda.
\end{align}
$x_\text{max}, x_\text{min}, y_\text{max}, y_\text{min}$ are the minimum and maximum values of $x_t, y_t$ from the supervised stroke labels and not the generated strokes. $W$ and $H$ are the width and height in pixels of our output canvas.

\paragraph{Calculate pixel intensity.}
Finally we are able to calculate the pixel $p_{ij}$ intensity of every pixel in our $H \times W$ canvas.
\begin{align}
    p_{ij} &= \sigma\Bigg[2 - 5 \times \min_{t=1\dots N_s} \bigg(\\&\text{dist}\big((i,j), (x_{t-1}, y_{t-1}), (x_t, y_t)\big) + (1 - \lfloor s_{1,t-1} \rceil) 10^6\bigg)\Bigg],
\end{align}
where the distance function is the distance between point $(i, j)$ from the line segment defined by the absolute points $(x_{t-1}, y_{t-1})$ and $(x_t, y_t)$. We also blow up any distances where $s_{1,t-1} < 0.5$ so as to not render any strokes where the pen is not touching the paper.
\vspace{-0.1in}
\section{Implementation Details}
\label{appendix:implementation_details}
We train our model for 300k iterations with a batch size of 256 for the Quickdraw dataset and 64 for Sketchy due to memory constraints. The initial learning rate is 1e-3 which decays by $0.85$ every 15k steps. We use the Adam~\citep{adam} optimizer and clip gradient values at $1.0$. $\sigma=2.0$ is used for the Gaussian blur in $\mathcal{L}_\text{pixel}$. For the curriculum learning schedule, the value of $\alpha$ is set to $0$ initially and increases by $0.05$ every 10k training steps with an empirically obtained cap at $\alpha_\text{max}=0.50$ for Quickdraw and $\alpha_\text{max}=0.75$ for Sketchy.

The ResNet12~\citep{oreshkin2018tadam} encoder uses 4 ResNet blocks with 64, 128, 256, 512 filters respectively and ReLU activations. The Conv4 backbone has 4 blocks of convolution, batch norm~\citep{batchnorm}, ReLU and max pool, identical to~\citet{vinyals2016matching}.  We select the latent space to be 256 dimensions, RNN output size to be 1024, and the hypernetwork embedding size to be 64. We use a mixture of $M = 30$ bivariate Gaussians for the mixture density output of the stroke offset distribution.
\section{Data Processing}
\label{appendix:data_processing}
\subsection{Quickdraw}
We apply the same data processing methods as in \citet{ha2017sketchrnn} with no additional changes to produce our stroke labels $\bm{y}$. When rasterizing for our input $\bm{x}$, we scale, center the strokes then pad the image with 10\% of the resolution in that dimension rounded to the nearest integer.

The following list of classes were used for training:
{\tiny 
The Eiffel Tower, The Mona Lisa, aircraft carrier, alarm clock, ambulance, angel, animal migration, ant, apple, arm, asparagus, banana, barn, baseball, baseball bat, bathtub, beach, bear, bed, bee, belt, bench, bicycle, binoculars, bird, blueberry, book, boomerang, bottlecap, bread, bridge, broccoli, broom, bucket, bulldozer, bus, bush, butterfly, cactus, cake, calculator, calendar, camel, camera, camouflage, campfire, candle, cannon, car, carrot, castle, cat, ceiling fan, cell phone, cello, chair, chandelier, church, circle, clarinet, clock, coffee cup, computer, cookie, couch, cow, crayon, crocodile, crown, cruise ship, diamond, dishwasher, diving board, dog, dolphin, donut, door, dragon, dresser, drill, drums, duck, dumbbell, ear, eye, eyeglasses, face, fan, feather, fence, finger, fire hydrant, fireplace, firetruck, fish, flamingo, flashlight, flip flops, flower, foot, fork, frog, frying pan, garden, garden hose, giraffe, goatee, grapes, grass, guitar, hamburger, hand, harp, hat, headphones, hedgehog, helicopter, helmet, hockey puck, hockey stick, horse, hospital, hot air balloon, hot dog, hourglass, house, house plant, ice cream, key, keyboard, knee, knife, ladder, lantern, leaf, leg, light bulb, lighter, lighthouse, lightning, line, lipstick, lobster, mailbox, map, marker, matches, megaphone, mermaid, microphone, microwave, monkey, mosquito, motorbike, mountain, mouse, moustache, mouth, mushroom, nail, necklace, nose, octopus, onion, oven, owl, paint can, paintbrush, palm tree, parachute, passport, peanut, pear, pencil, penguin, piano, pickup truck, pig, pineapple, pliers, police car, pool, popsicle, postcard, purse, rabbit, raccoon, radio, rain, rainbow, rake, remote control, rhinoceros, river, rollerskates, sailboat, sandwich, saxophone, scissors, see saw, shark, sheep, shoe, shorts, shovel, sink, skull, sleeping bag, smiley face, snail, snake, snowflake, soccer ball, speedboat, square, star, steak, stereo, stitches, stop sign, strawberry, streetlight, string bean, submarine, sun, swing set, syringe, t-shirt, table, teapot, teddy-bear, tennis racquet, tent, tiger, toe, tooth, toothpaste, tractor, traffic light, train, triangle, trombone, truck, trumpet, umbrella, underwear, van, vase, watermelon, wheel, windmill, wine bottle, wine glass, wristwatch, zigzag, blackberry, power outlet, peas, hot tub, toothbrush, skateboard, cloud, elbow, bat, pond, compass, elephant, hurricane, jail, school bus, skyscraper, tornado, picture frame, lollipop, spoon, saw, cup, roller coaster, pants, jacket, rifle, yoga, toilet, waterslide, axe, snowman, bracelet, basket, anvil, octagon, washing machine, tree, television, bowtie, sweater, backpack, zebra, suitcase, stairs, The Great Wall of China
}

\subsection{Omniglot}
We derive our Omniglot tasks from the stroke dataset originally provided by \citet{Lake2015bpl} rather than the image analogues. We translate the Omniglot stroke-by-stroke format to the same one used in Quickdraw. Then we apply the Ramer-Douglas-Peucker \citep{Douglas1973rdp} algorithm with an epsilon value of 2 and normalize variance to $1$ to produce $\bm{y}$. We also rasterize our images in the same manner as above for our input $\bm{x}$.

\subsection{Sketchy}
Sketchy data is provided as an SVG image composed of line paths that are either straight lines or Bezier curves. To generate stroke data we sample sequences of points from Bezier curves at a high resolution that we then simplify with RDP, $\epsilon=5$. We also eliminate continuous strokes with a short path length or small displacement to reduce our stroke length and remove small and noisy strokes. Path length and displacement are considered with respect to the scale of the entire sketch.

Once again we normalize stroke variance and rasterize for our input image in the same manners as above.

The following classes were use for training after removing overlapping classes with mini-ImageNet:
{\tiny 
hot-air\_balloon, violin, tiger, eyeglasses, mouse, jack-o-lantern, lobster, teddy\_bear, teapot, helicopter, duck, wading\_bird, rabbit, penguin, sheep, windmill, piano, jellyfish, table, fan, beetle, cabin, scorpion, scissors, banana, tank, umbrella, crocodilian, volcano, knife, cup, saxophone, pistol, swan, chicken, sword, seal, alarm\_clock, rocket, bicycle, owl, squirrel, hermit\_crab, horse, spoon, cow, hotdog, camel, turtle, pizza, spider, songbird, rifle, chair, starfish, tree, airplane, bread, bench, harp, seagull, blimp, apple, geyser, trumpet, frog, lizard, axe, sea\_turtle, pretzel, snail, butterfly, bear, ray, wine\_bottle, , elephant, raccoon, rhinoceros, door, hat, deer, snake, ape, flower, car\_(sedan), kangaroo, dolphin, hamburger, castle, pineapple, saw, zebra, candle, cannon, racket, church, fish, mushroom, strawberry, window, sailboat, hourglass, cat, shoe, hedgehog, couch, giraffe, hammer, motorcycle, shark
}

\section{Pixel-loss Weighting $\alpha_\text{max}$ Ablation for Generation Quality}
\begin{table}[h]
\begin{center}
    \caption{Effect of $\alpha_\text{max}$ on classification accuracy of generated sketches.}
    \label{tab:alpha_sweep_generation}
    \resizebox{0.48\textwidth}{!}{\begin{tabular}{c|cccccc}
        $\alpha_\text{max}$ & 0.00 & 0.25 & 0.50 & 0.75 & 0.95 & 1.00     \\ \midrule
        Seen & 87.76 & 87.35 & 81.44 & 66.80 & 36.98 & 04.80 \\
        Unseen & 84.02 & 85.32 & 77.94 & 63.10 & 32.94 & 04.50 \\
    \end{tabular}}
\end{center}
\vspace{-0.1in}
\end{table}

\begin{figure}[h]
        \centering
     \begin{subfigure}{0.48\textwidth}
      \caption{Autoregressive generation.} 
      \vspace{-0.1in}
         \centering
         \includegraphics[trim=0.4cm 14.3cm 24.8cm 0.35cm,clip,width=0.8\textwidth]{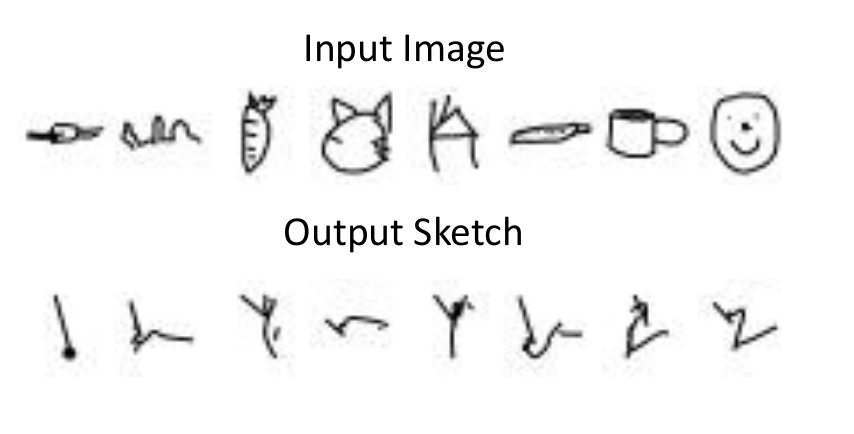}

     \end{subfigure}
     \vspace{0.3in}
    \begin{subfigure}{0.48\textwidth}
             \caption{Teacher-forced generation.}
             \vspace{-0.1in}
         \centering
         \includegraphics[trim=0.4cm 15.8cm 25.3cm 0.2cm,clip,width=0.8\textwidth]{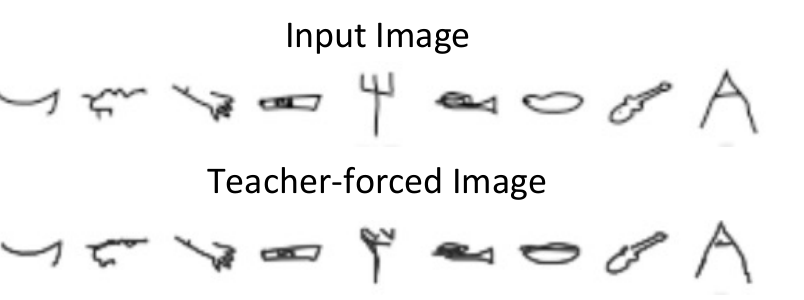}
     \end{subfigure}
     \caption{Sketches of \model{} trained with $\alpha_\text{max}=1.0$.}
     \label{fig:alphamax1}
\end{figure}
We also ablate the impact of pixel-loss weighting parameter $\alpha_\text{max}$ on the classification accuracy of the ResNet models from Section~\ref{sec:generation_quality}. The evaluation process is the same, generating sketches of examples from classes that were either seen during training or new to the model and classifying them in 45-way classification. Results are shown in Table~\ref{tab:alpha_sweep_generation}. 

Results are only shown for the Quickdraw~\cite{jongejan2016quickdraw} setting. Increasing pixel-loss weighting has a minor impact on classification accuracy at lower values but has a significant detriment at higher weightings. This is due to the teacher-forcing training process. As we de-weight the 
stroke loss, the model no longer learns to handle the uncertainty of the input position in the space of the 2D canvas by predicting a distribution that explains the next ground truth point. It only matches the generation in pixel space and no longer generates a sensible stroke trajectory on the canvas. While training under teacher forcing, this is not an issue as it is fed the ground truth input point every time, but in autoregressive this generation quickly degrades as each step no longer produces the a point that is a meaningful input for the next time step. We can see the significant difference between generation quality under techer forcing and autoregressive generation in Figure~\ref{fig:alphamax1}.


\begin{figure*}
    \centering
    \begin{subfigure}{0.48\textwidth}
         \centering
         \includegraphics[width=\textwidth]{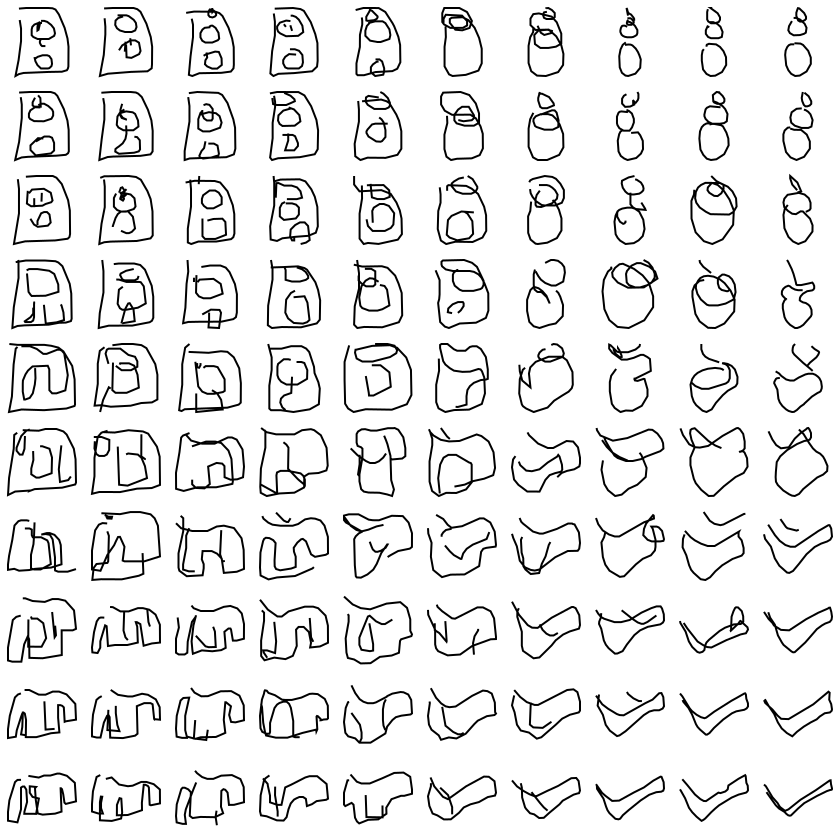}
         \caption{Interpolation of classes: power outlet, snowman, jacket, elbow.}
         \label{fig:interp1}
     \end{subfigure}
     \hfill
     \begin{subfigure}{0.48\textwidth}
         \centering
         \includegraphics[width=\textwidth]{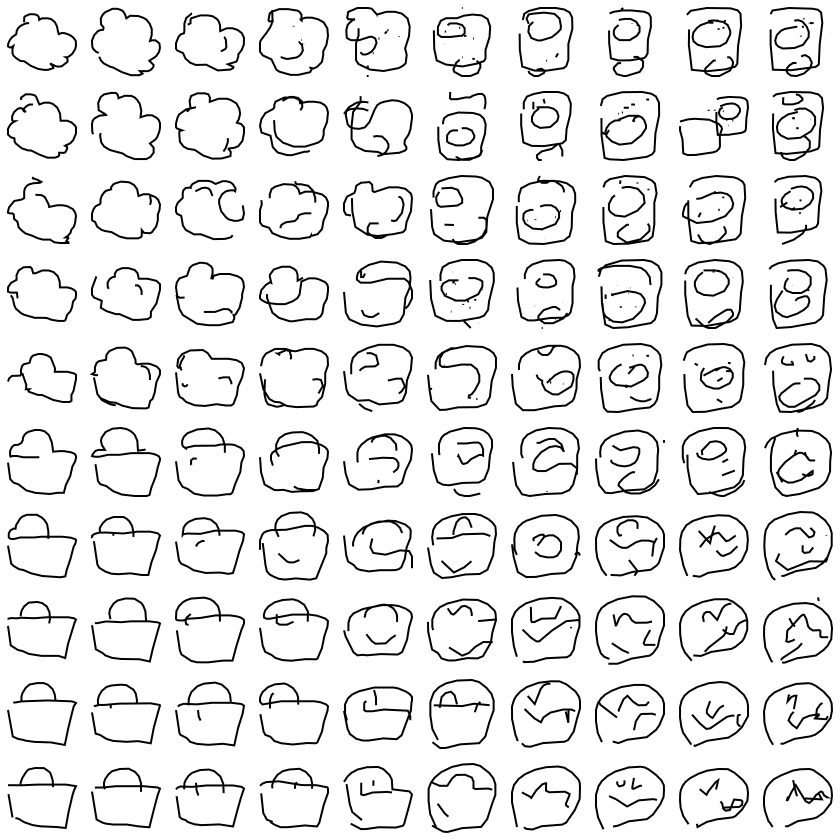}
         \caption{Interpolation of classes: cloud, power outlet, basket, compass.}
         \label{fig:interp2}
     \end{subfigure}
     \caption{Latent space interpolations of randomly selected examples.}
\end{figure*}

\section{Latent Space Interpolation}
\label{appendix:interpolation}
Like in many encoding-decoding models we evaluate the interpolation of our latent space. We select 4 embeddings at random and use bi-linear interpolation to produce new embeddings. Results are in Figures \ref{fig:interp1} and \ref{fig:interp2}.

We observe that compositionality is also present in these interpolations. In the top row of Figure \ref{fig:interp1}, the model first plots a third small circle when interpolating from the 2-circle power outlet and the 3-circle snowman. This small circle is treated as single component that grows as it transitions between classes until it's final size in the far right snowman drawing.

Some other RNN-based sketching models \citep{ha2017sketchrnn, chen2017pix2seq} experience other classes materializing in interpolations between two unrelated classes. Our model does not exhibit this same behaviour as our embedding space is learned from more classes and thus does not contain local groupings of classes.
\section{Intra-alphabet Lake Split}
\label{appendix:lake_split}
The creators of the Omniglot dataset and one-shot classification benchmark originally proposed an intra-alphabet classification task. This task is more challenging than the common Vinyals split as characters from the same alphabet may exhibit similar stylistics of sub-components that makes visual differentiation more difficult. This benchmark has been less explored by researchers; however, we still present the performance of our \modelembedding{}-based approach against other few-shot classification models on the benchmark. Results are shown in Table \ref{tab:lakesplit_results}. 

\begin{table*}
    \centering
    \caption{Few-shot classification results on Omniglot (Lake split).}
    \resizebox{0.98\textwidth}{!}{
    \begin{tabular}{@{}lllllll@{}}
\toprule
\multicolumn{3}{c}{\textbf{Omniglot (Lake split)}}     & \multicolumn{4}{c}{\textbf{(way, shot)}}         \\ \midrule
\textbf{Algorithm}        & \textbf{Backbone} & \textbf{Train Data} & \multicolumn{1}{c}{\textbf{(5,1)}} & \multicolumn{1}{c}{\textbf{(5,5)}} & \multicolumn{1}{c}{\textbf{(20,1)}} & \multicolumn{1}{c}{\textbf{(20,5)}}        \\ \midrule
Conv-VAE                                      & Conv4 & Quickdraw           & 73.12 $\pm$ 0.58 & 88.50 $\pm$ 0.39 & 53.45 $\pm$ 0.51 & 73.62 $\pm$ 0.48  \\ \midrule
\model{} \textit{(Ours)}               & Conv4 & Quickdraw           & 89.16 $\pm$ 0.41 & 97.12 $\pm$ 0.18 & 74.24 $\pm$ 0.48 & 89.87 $\pm$ 0.25  \\ 
\model{} \textit{(Ours)}               & ResNet12 & Quickdraw        & \textbf{91.03} $\pm$ 0.37 & \textbf{97.91} $\pm$ 0.15   & \textbf{77.94} $\pm$ 0.44   & \textbf{92.49} $\pm$ 0.21    \\ \midrule
BPL \textit{(Supervised)} \citep{Lake2015bpl,Lake2019omnifollowup}  & N/A   & Omniglot            & \multicolumn{1}{c}{-} & \multicolumn{1}{c}{-} & \multicolumn{1}{c}{96.70} & \multicolumn{1}{c}{-} \\
ProtoNet \textit{(Supervised)} \citep{snell2017prototypical,Lake2019omnifollowup}                   & Conv4 & Omniglot            & \multicolumn{1}{c}{-} & \multicolumn{1}{c}{-} & \multicolumn{1}{c}{86.30} & \multicolumn{1}{c}{-} \\
RCN \textit{(Supervised)} \citep{rcn,Lake2019omnifollowup} & N/A   & Omniglot            & \multicolumn{1}{c}{-} & \multicolumn{1}{c}{-} & \multicolumn{1}{c}{92.70} & \multicolumn{1}{c}{-} \\
VHE \textit{(Supervised)} \citep{vhe,Lake2019omnifollowup}   & N/A   & Omniglot            & \multicolumn{1}{c}{-} & \multicolumn{1}{c}{-} & \multicolumn{1}{c}{81.30} & \multicolumn{1}{c}{-} \\ \bottomrule
    \end{tabular}}
    \label{tab:lakesplit_results}
\end{table*}

Unsurprisingly, our model is outperformed by supervised models and does fall behind by a more substantial margin than in the Vinyals split. However, our method approach still achieves respectable classification accuracy overall and greatly outperforms a Conv-VAE baseline.
\section{Effect of Random Seeding on Few-Shot Classification}
\label{appendix:seeding}
The training objective for \model is to reproduce sketch drawings of the input. This task is unrelated to few-shot classification may perform variably given different initialization. We quantify this variance by training our model with 15 unique random seeds and evaluating the performance of the latent space on the few-shot classification tasks.

We disregard the per (evaluation) episode variance of our model in each test stage and only present the mean accuracy. We then compute a new confidence interval over random seeds. Results are presented in Tables \ref{tab:omniglot_conv4_seed_results}, \ref{tab:mini-imagenet_seed_results}.

\begin{table*}
    \caption{Few-shot classification random seeding experiments.}
    \centering
    \begin{subtable}[h]{0.48\textwidth}
    \caption{Omniglot (Conv4).}
    \resizebox{0.98\textwidth}{!}{
    \begin{tabular}{@{} cccccc @{}}
    \toprule
    & \multicolumn{4}{c}{\textbf{(way, shot)}}         \\ \midrule
    \textbf{Seed} & \textbf{(5,1)} & \textbf{(5,5)} & \textbf{(20,1)} & \textbf{(20,5)}        \\ \midrule
    1       & 96.45 & 99.41 & 90.84 & 98.08 \\
    2       & 96.54 & 99.48 & 90.82 & 98.10 \\
    3       & 96.23 & 99.40 & 90.05 & 97.94 \\
    4       & 96.15 & 99.46 & 90.50 & 97.99 \\
    5       & 96.21 & 99.40 & 90.54 & 98.10 \\
    6       & 96.08 & 99.43 & 90.20 & 97.93 \\
    7       & 96.19 & 99.39 & 90.70 & 98.05 \\
    8       & 96.68 & 99.44 & 91.11 & 98.18 \\
    9       & 96.49 & 99.42 & 90.64 & 98.06 \\
    10      & 96.37 & 99.47 & 90.50 & 97.99 \\
    11      & 96.52 & 99.40 & 91.13 & 98.18 \\
    12      &\textbf{ 96.96} & \textbf{99.50} & \textbf{91.67} & \textbf{98.30} \\
    13      & 96.31 & 99.38 & 90.57 & 98.04 \\
    14      & 96.12 & 99.45 & 90.54 & 98.03 \\
    15      & 96.30 & 99.48 & 90.62 & 98.05 \\\midrule 
    Average & 96.37 $\pm$ 0.12 & 99.43 $\pm$ 0.02 & 90.69 $\pm$ 0.20 & 98.07 $\pm$ 0.05\\ \bottomrule
    \end{tabular}}
    \label{tab:omniglot_conv4_seed_results}
    \end{subtable}
    \centering
    \begin{subtable}[h]{0.48\textwidth}
    \caption{mini-ImageNet.}
    \resizebox{0.98\textwidth}{!}{
    \begin{tabular}{@{}cccccc@{}}
\toprule
& \multicolumn{4}{c}{\textbf{(way, shot)}}         \\ \midrule
\textbf{Seed} & \textbf{(5,1)} & \textbf{(5,5)} & \textbf{(5,20)} & \textbf{(5,50)}        \\ \midrule
1       & 37.15 & 52.99 & 63.92 & 68.72 \\
2       & 39.38 & 55.20 & 65.60 & 69.79 \\
3       & 39.40 & 55.47 & 65.94 & 70.41 \\
4       & \textbf{40.39} & \textbf{57.15} & \textbf{67.60} & \textbf{71.99} \\
5       & 38.40 & 54.08 & 65.36 & 70.08 \\
6       & 37.94 & 53.98 & 65.24 & 69.65 \\
7       & 38.88 & 55.71 & 66.59 & 71.35 \\
8       & 37.89 & 52.65 & 63.42 & 68.14 \\
9       & 38.25 & 53.86 & 65.02 & 69.82 \\
10      & 39.11 & 55.29 & 65.99 & 69.98 \\
11      & 37.39 & 52.88 & 63.66 & 68.33 \\
12      & 38.24 & 53.91 & 65.19 & 69.82 \\
13      & 38.62 & 53.84 & 63.83 & 68.69 \\
14      & 37.73 & 53.61 & 64.22 & 68.41 \\
15      & 39.50 & 55.23 & 65.51 & 70.25 \\\midrule 
Average & 38.55 $\pm$ 0.45 & 54.39 $\pm$ 0.63 & 65.14 $\pm$ 0.59 & 69.69 $\pm$ 0.56\\ \bottomrule
    \end{tabular}
    }
    \label{tab:mini-imagenet_seed_results}
    \end{subtable}
\end{table*}

\section{Few-shot Classification on Omniglot -- Full Results.}
\label{appendix:full_omniglot_table}
The full results (Table \ref{tab:omniglot_results_full}) for few-shot classification on the Omniglot \citep{Lake2015bpl} dataset, including the ResNet12 \citep{oreshkin2018tadam} model. We provide results on SketchEmbedNet trained with a KL objective on the latent representation. The (\textit{w/ Labels}) is a model variant where there is an additional head predicting the class from the latent representation while sketching the class. This was to hopefully learn a more discriminative embedding, except it lowered classification accuracy.
\begin{table*}
    \centering
    \caption{Full table of few-shot classification results on Omniglot.}
    \resizebox{0.98\textwidth}{!}{
    \begin{threeparttable}
    \begin{tabular}{@{}lllllll@{}}
\toprule
\multicolumn{3}{c}{\textbf{Omniglot}}     & \multicolumn{4}{c}{\textbf{(way, shot)}}         \\ \midrule
\textbf{Algorithm}        & \textbf{Backbone} & \textbf{Train Data} & \multicolumn{1}{c}{\textbf{(5,1)}} & \multicolumn{1}{c}{\textbf{(5,5)}} & \multicolumn{1}{c}{\textbf{(20,1)}} & \multicolumn{1}{c}{\textbf{(20,5)}}        \\ \midrule
Training from Scratch    \citep{hsu2018cactus} & N/A & Omniglot            & 52.50 $\pm$ 0.84 & 74.78 $\pm$ 0.69 & 24.91 $\pm$ 0.33 & 47.62 $\pm$ 0.44  \\ \midrule
Random CNN                                    & Conv4 & N/A                 & 67.96 $\pm$ 0.44 & 83.85 $\pm$ 0.31 & 44.39 $\pm$ 0.23 & 60.87 $\pm$ 0.22  \\
Conv-VAE                                      & Conv4 & Omniglot            & 77.83 $\pm$ 0.41 & 92.91 $\pm$ 0.19 & 62.59 $\pm$ 0.24 & 84.01 $\pm$ 0.15  \\
Conv-VAE                                      & Conv4 & Quickdraw           & 81.49 $\pm$ 0.39 & 94.09 $\pm$ 0.17 & 66.24 $\pm$ 0.23 & 86.02 $\pm$ 0.14  \\ 
Conv-AE                                       & Conv4 & Quickdraw           & 81.54 $\pm$ 0.40 & 93.57 $\pm$ 0.19 & 67.24 $\pm$ 0.24 & 84.15 $\pm$ 0.16  \\
$\beta$-VAE ($\beta=250$)     \citep{betavae} & Conv4 & Quickdraw           & 79.11 $\pm$ 0.40 & 93.23 $\pm$ 0.19 & 63.67 $\pm$ 0.24 & 84.92 $\pm$ 0.15  \\ 
k-NN                     \citep{hsu2018cactus} & N/A & Omniglot            & 57.46 $\pm$ 1.35 & 81.16 $\pm$ 0.57 & 39.73 $\pm$ 0.38 & 66.38 $\pm$ 0.36  \\
Linear Classifier        \citep{hsu2018cactus} & N/A & Omniglot            & 61.08 $\pm$ 1.32 & 81.82 $\pm$ 0.58 & 43.20 $\pm$ 0.69 & 66.33 $\pm$ 0.36  \\
MLP + Dropout            \citep{hsu2018cactus} & N/A & Omniglot            & 51.95 $\pm$ 0.82 & 77.20 $\pm$ 0.65 & 30.65 $\pm$ 0.39 & 58.62 $\pm$ 0.41  \\
Cluster Matching         \citep{hsu2018cactus} & N/A & Omniglot            & 54.94 $\pm$ 0.85 & 71.09 $\pm$ 0.77 & 32.19 $\pm$ 0.40 & 45.93 $\pm$ 0.40  \\
CACTUs-MAML              \citep{hsu2018cactus} & Conv4 & Omniglot            & 68.84 $\pm$ 0.80 & 87.78 $\pm$ 0.50 & 48.09 $\pm$ 0.41 & 73.36 $\pm$ 0.34  \\
CACTUs-ProtoNet          \citep{hsu2018cactus} & Conv4 & Omniglot            & 68.12 $\pm$ 0.84 & 83.58 $\pm$ 0.61 & 47.75 $\pm$ 0.43 & 66.27 $\pm$ 0.37  \\
AAL-ProtoNet        \citep{antoniou2019assume} & Conv4 & Omniglot            & 84.66 $\pm$ 0.70 & 88.41 $\pm$ 0.27 & 68.79 $\pm$ 1.03 & 74.05 $\pm$ 0.46  \\
AAL-MAML            \citep{antoniou2019assume} & Conv4 & Omniglot            & 88.40 $\pm$ 0.75 & 98.00 $\pm$ 0.32 & 70.20 $\pm$ 0.86 & 88.30 $\pm$ 1.22  \\
UMTRA              \citep{khodadadeh2019umtra} & Conv4 & Omniglot            & 83.80            & 95.43            & 74.25            & 92.12             \\ \midrule
Contrastive                                    & Conv4 & Omniglot*           & 77.69 $\pm$ 0.40 & 92.62 $\pm$ 0.20 & 62.99 $\pm$ 0.25 & 83.70 $\pm$ 0.16 \\
\model{} \textit{(Ours)}               & Conv4 & Omniglot*            & 94.88 $\pm$ 0.22 & 99.01 $\pm$ 0.08 & 86.18 $\pm$ 0.18 & 96.69 $\pm$ 0.07  \\
Contrastive                                    & Conv4 & Quickdraw*          & 83.26 $\pm$ 0.40 & 94.16 $\pm$ 0.21 & 73.01 $\pm$ 0.25 & 86.66 $\pm$ 0.17  \\
\model{}-avg  \textit{(Ours)}               & Conv4 & Quickdraw*           & 96.37            & 99.43            & 90.69            & 98.07 \\ 
\model{}-best \textit{(Ours)}               & Conv4 & Quickdraw*           & \textbf{96.96} $\pm$ 0.17 & \textbf{99.50} $\pm$ 0.06 & \textbf{91.67} $\pm$ 0.14 & 98.30 $\pm$ 0.05  \\ 
\model{}-avg  \textit{(Ours)}               & ResNet12 & Quickdraw*           & 96.00            & 99.51                       & 89.88              & 98.27 \\ 
\model{}-best \textit{(Ours)}               & ResNet12 & Quickdraw*           & 96.61 $\pm$ 0.19 & \textbf{99.58} $\pm$ 0.06   & 91.25 $\pm$ 0.15   & \textbf{98.58} $\pm$ 0.05    \\ \midrule
\model{}(KL)-avg  \textit{(Ours)}               & Conv4 & Quickdraw*           & 96.06            & 99.40            & 89.83            & 97.92 \\ 
\model{}(KL)-best \textit{(Ours)}                 & Conv4 & Quickdraw*           & 96.60 $\pm$ 0.18 & 99.46 $\pm$ 0.06 & 90.84 $\pm$ 0.15 & 98.09 $\pm$ 0.06 \\ \midrule
\model{} \textit{(w/ Labels)} \textit{(Ours)}     & Conv4 & Quickdraw*           & 88.52 $\pm$ 0.34 & 96.73 $\pm$ 0.13 & 71.35 $\pm$ 0.24 & 88.16 $\pm$ 0.14 \\ \midrule
MAML \textit{(Supervised)} \citep{finn2017maml}                 & Conv4 & Omniglot            & 94.46 $\pm$ 0.35 & 98.83 $\pm$ 0.12 & 84.60 $\pm$ 0.32 & 96.29 $\pm$ 0.13  \\
ProtoNet \textit{(Supervised)} \citep{snell2017prototypical}    & Conv4 & Omniglot            & 98.35 $\pm$ 0.22 & 99.58 $\pm$ 0.09 & 95.31 $\pm$ 0.18 & 98.81 $\pm$ 0.07  \\ \bottomrule
    \end{tabular}
    \begin{tablenotes}\footnotesize
    \item[*] Sequential sketch supervision used for training
    \end{tablenotes}
    \end{threeparttable}}
    \label{tab:omniglot_results_full}
\end{table*}

\section{Few-shot Classification on mini-ImageNet -- Full Results}
\label{appendix:full_mii_table}
The full results (Table \ref{tab:mini-imagenet_results_full}) for few-shot classification on the mini-ImageNet dataset, including the ResNet12 \citep{oreshkin2018tadam} model and Conv4 models.

\begin{table*}
    \caption{Full table of few-shot classification results on mini-ImageNet.}
    \centering
    \resizebox{0.98\textwidth}{!}{
    \begin{threeparttable}
    \begin{tabular}{@{}lllllll@{}}
\toprule
\multicolumn{3}{c}{\textbf{mini-ImageNet}}& \multicolumn{4}{c}{\textbf{(way, shot)}}         \\ \midrule
\textbf{Algorithm}       & \textbf{Backbone} & \textbf{Train Data} & \multicolumn{1}{c}{\textbf{(5,1)}} & \multicolumn{1}{c}{\textbf{(5,5)}} & \multicolumn{1}{c}{\textbf{(5,20)}} & \multicolumn{1}{c}{\textbf{(5,50)}}        \\ \midrule
Training from Scratch   \citep{hsu2018cactus} & N/A & mini-ImageNet       & 27.59 $\pm$ 0.59 & 38.48 $\pm$ 0.66 & 51.53 $\pm$ 0.72 & 59.63 $\pm$ 0.74  \\ \midrule
UMTRA             \citep{khodadadeh2019umtra} & Conv4 & mini-ImageNet       & 39.93   & 50.73            & 61.11            & 67.15             \\
CACTUs-MAML              \citep{hsu2018cactus} & Conv4 & mini-ImageNet       & 39.90 $\pm$ 0.74 & 53.97 $\pm$ 0.70 & 63.84 $\pm$ 0.70 & 69.64 $\pm$ 0.63  \\
CACTUs-ProtoNet          \citep{hsu2018cactus} & Conv4 & mini-ImageNet       & 39.18 $\pm$ 0.71 & 53.36 $\pm$ 0.70 & 61.54 $\pm$ 0.68 & 63.55 $\pm$ 0.64  \\
AAL-ProtoNet       \citep{antoniou2019assume} & Conv4 & mini-ImageNet       & 37.67 $\pm$ 0.39 & 40.29 $\pm$ 0.68 &\multicolumn{1}{c}{-}&\multicolumn{1}{c}{-} \\
AAL-MAML           \citep{antoniou2019assume} & Conv4 & mini-ImageNet       & 34.57 $\pm$ 0.74 & 49.18 $\pm$ 0.47 &\multicolumn{1}{c}{-}&\multicolumn{1}{c}{-} \\
Random CNN                                   & Conv4 & N/A       & 26.85 $\pm$ 0.31 & 33.37 $\pm$ 0.32 & 38.51 $\pm$ 0.28 & 41.41 $\pm$ 0.28  \\ 
Conv-VAE                                     & Conv4 & mini-ImageNet       & 23.30 $\pm$ 0.21 & 26.22 $\pm$ 0.20 & 29.93 $\pm$ 0.21 & 32.57 $\pm$ 0.20  \\
Conv-VAE                                     & Conv4 & Sketchy       & 23.27 $\pm$ 0.18 & 26.28 $\pm$ 0.19 & 30.41 $\pm$ 0.19 & 33.97 $\pm$ 0.19  \\ 
Random CNN                                   & ResNet12 & N/A                & 28.59 $\pm$ 0.34 & 35.91 $\pm$ 0.34 & 41.31 $\pm$ 0.33 & 44.07 $\pm$ 0.31  \\ 
Conv-VAE                                     & ResNet12 & mini-ImageNet      & 23.82 $\pm$ 0.23 & 28.16 $\pm$ 0.25 & 33.64 $\pm$ 0.27 & 37.81 $\pm$ 0.27  \\
Conv-VAE                                     & ResNet12 & Sketchy            & 24.61 $\pm$ 0.23 & 28.85 $\pm$ 0.23 & 35.72 $\pm$ 0.27 & 40.44 $\pm$ 0.28  \\ \midrule
Contrastive                  & ResNet12 & Sketchy*           & 30.56 $\pm$ 0.33 & 39.06 $\pm$ 0.33 & 45.17 $\pm$ 0.33 & 47.84 $\pm$ 0.32  \\ 
\model{}-avg \textit{(ours)}               & Conv4 & Sketchy*                & 37.01  & 51.49  & 61.41  & 65.75  \\
\model{}-best \textit{(ours)}              & Conv4 & Sketchy*                & 38.61 $\pm$ 0.42 & 53.82 $\pm$ 0.41 & 63.34 $\pm$ 0.35 & 67.22 $\pm$ 0.32 \\
\model{}-avg \textit{(ours)}               & ResNet12 & Sketchy*             & 38.55  & 54.39  & 65.14  & 69.70  \\
\model{}-best \textit{(ours)}              & ResNet12 & Sketchy*             & \textbf{40.39} $\pm$ 0.44 & \textbf{57.15} $\pm$ 0.38 & \textbf{67.60} $\pm$ 0.33 & \textbf{71.99} $\pm$ 0.3 \\\midrule
MAML \textit{(supervised)} \citep{finn2017maml}                 & Conv4 & mini-ImageNet       & 46.81 $\pm$ 0.77 & 62.13 $\pm$ 0.72 & 71.03 $\pm$ 0.69 & 75.54 $\pm$ 0.62  \\
ProtoNet \textit{(supervised)} \citep{snell2017prototypical}    & Conv4 & mini-ImageNet       & 46.56 $\pm$ 0.76 & 62.29 $\pm$ 0.71 & 70.05 $\pm$ 0.65 & 72.04 $\pm$ 0.60  \\ \bottomrule
    \end{tabular}
    \begin{tablenotes}\footnotesize
    \item[*] Sequential sketch supervision used for training
    \end{tablenotes}
    \end{threeparttable}}
    \label{tab:mini-imagenet_results_full}
\end{table*}

\clearpage

\end{document}